\documentclass[journal]{IEEEtran}
\usepackage{blindtext}
\usepackage{graphicx}
\usepackage{multirow}
\usepackage{booktabs}
\usepackage{comment}
\usepackage{xcolor}
\usepackage{amssymb}
\usepackage{mathtools}
\usepackage{hyperref}
\usepackage{tikz}
\usepackage{pgfplots}
\usepackage{color, colortbl}
\usepackage{amsmath}
\usepackage{url}

\begin{document}

\title{Deep Relation Learning for Regression and Its Application to Brain Age Estimation}

\author{Sheng He, Yanfang Feng, P. Ellen Grant, Yangming Ou
       
\thanks{
Manuscript received by XXXX, accted XXXX. The work of Sheng He was supported by Charles A. King Trust Research Fellowship. The work of Yangming Ou was supported in part by Harvard Medical School/Boston Children's Hospital Faculty Development Award, and in part by St. Baldrick Foundation Scholar Award Grace Fund and R03 HD104891, R21 NS121735.
The work of Ellen Grant was partly supported by R01HL152358, R01EB32708, U01DA055353, R01HD065762, and R01HD100009.
(\textit{Sheng He and Yanfang Feng are co-first authors.  Yangming Ou is the corresponding author.}) \\
S. He, P. Grant and Y. Ou are with the Boston Children's Hospital and Harvard Medical School, Harvard University, 300 Longwood Ave., Boston, MA, USA.
Y. Feng is with the Massachusetts General Hospital and Harvard Medical School, Harvard University, 55 Fruit St., Boston, MA, USA.\\
E-mail: sheng.he@childrens.harvard.edu; yfeng0@mgh.harvard.edu,
ellen.grant@childrens.harvard.edu, yangming.ou@childrens.harvard.edu}
}

\maketitle

\begin{abstract}
Most deep learning models for temporal regression directly output the estimation based on single input images, ignoring the relationships between different images.
In this paper, we propose deep relation learning for regression, aiming to learn different relations between a pair of input images.
Four non-linear relations are considered: ``cumulative relation", ``relative relation", ``maximal relation" and ``minimal relation".
These four relations are learned simultaneously from one deep neural network which has two parts: feature extraction and relation regression.
We use an efficient convolutional neural network to extract deep features from the pair of input images and apply a Transformer for relation learning.
The proposed method is evaluated on a merged dataset with 6,049 subjects with ages of 0-97 years using 5-fold cross-validation for the task of brain age estimation.
The experimental results have shown that the proposed method achieved a mean absolute error (MAE) of 2.38 years, which is lower than the MAEs of 8 other state-of-the-art algorithms with statistical significance (p$<$0.05) in paired T-test (two-side).
\end{abstract}


\IEEEpeerreviewmaketitle

\section{Introduction}
\label{sec:intro}

Regression aims to estimate continuous values (or ordinal outcomes~\cite{lim2019order}) from the input data using machine learning models. 
It has many applications such as severity scores~\cite{lu2021quantifying,signoroni2021bs}, 
brain age estimation~\cite{cole2017predicting,feng2020estimating} and fluid intelligence prediction~\cite{saha2021predicting}.
Deep convolutional neural networks (CNNs) can transform the raw input image data into target variables by training on a large-scale dataset~\cite{lecun2015deep}. 
Therefore, many recent applications use deep learning to solve the regression problem~\cite{feng2020estimating,jonsson2019brain,brainwaa160}.

\begin{figure}[!t]
    \centering
    \includegraphics[width=0.5\textwidth]{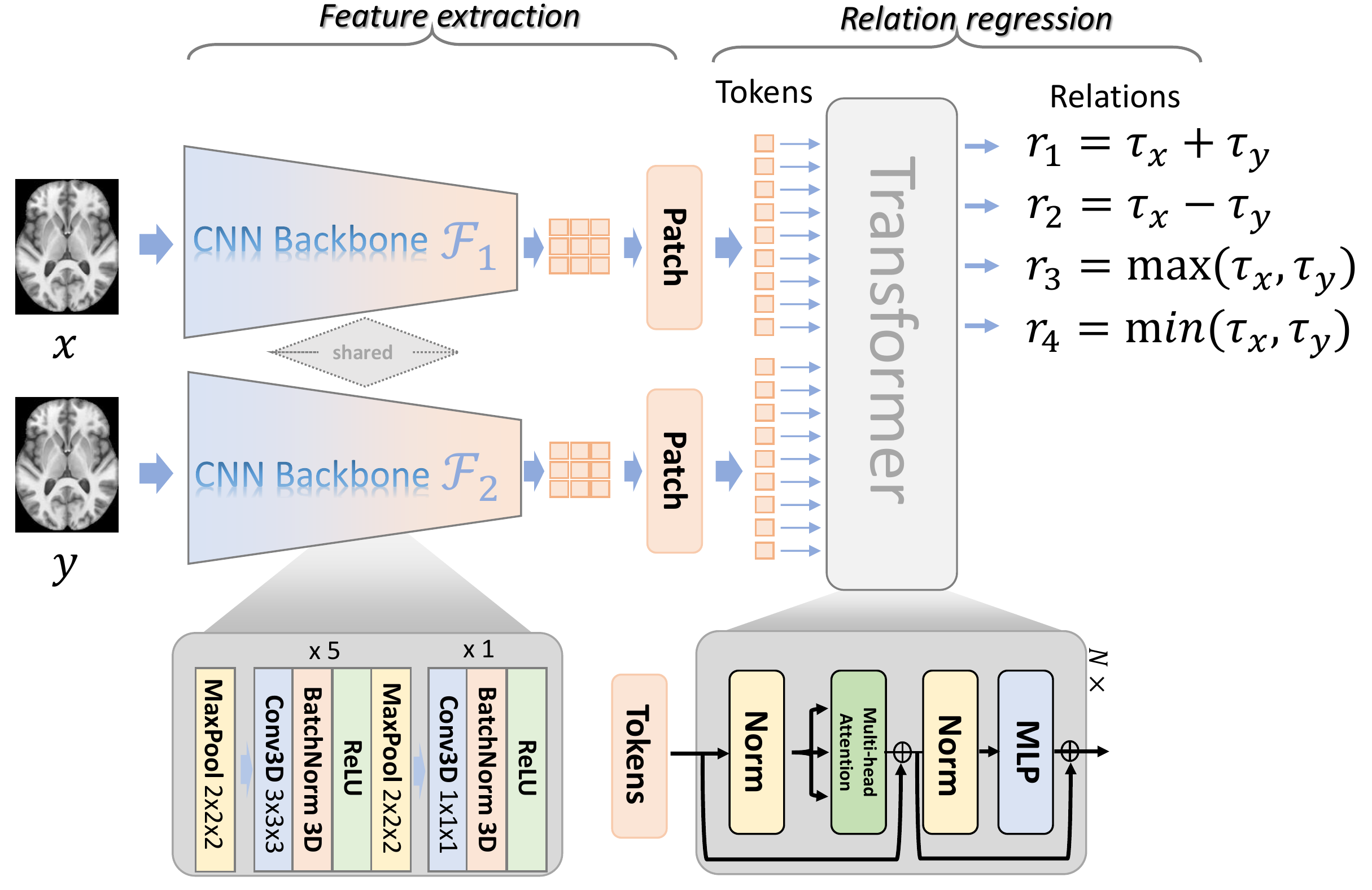}
    \caption{The framework of the proposed deep relation learning for brain age estimation has two parts: feature extraction and relation regression. It has the pair of inputs ($x,y)$ and the CNN backbones $\mathcal{F}_1$ and $\mathcal{F}_2$ for feature extraction. The two backbones can be shared (Siamese) or be independent, which stack several Convolutional, Batch Normalization, Max-Pooling and ReLU layers. The outputs of the CNN backbones are split into tokens to the standard transformer for relation regression: learning the four relations between the pair of inputs $x$ and $y$.}
    \label{fig:framework}
\end{figure}

Most deep regression methods estimate the ordinal output $z$ based on a single input image $x$ sampled from a set $\mathcal{X}$, which can be denoted by $z=\mathcal{F}_{\vec{w}}(x)$ (where $\mathcal{F}_{\vec{w}}$ is the machine learning model and ${\vec{w}}$ is the set of parameters).
This requires the machine learning model $\mathcal{F}_{\vec{w}}$ to learn the output directly from the single input image $x$ without any reference.
Instead of learning from the single image, pairwise learning $r=\mathcal{F}_{\vec{w}}(x,y)$ is also used in regression~\cite{lim2019order,zhang2017age,lee2019monocular} and segmentation~\cite{wang2021pairwise}, aiming to learn the ordinal relationship $r$ between the pair of input subjects $(x,y)$, such as the ternary relationship~\cite{lim2019order}: ``greater than", ``similar to" or ``smaller than".
The order regression based on pairwise learning has two advantages: 
(1) it is easy to learn the relationship between two instances~\cite{chen2016single} and (2) $y$ can be used as the reference to estimate 
an ordinal output of  $x$~\cite{lim2019order}.
However, limitations of the order learning with the ternary relationship proposed in~\cite{lim2019order} include: (1) it only learns one relationship between $x$ and $y$, thus the estimation of 
the ordinal output of 
$x$ needs a chain with many different references $y$ and (2) it lacks the reflexivity~\cite{wang2021pairwise} which cannot estimate the output with the pair of the identical input $(x,x)$.
(The reflexivity means the model can provide the estimation based on a pair of identical input images.)

 In this paper, we propose a novel deep relation learning framework to solve these limitations of ordinal learning.
Given two sets $\mathcal{X}$ and $\mathcal{Y}$, the Cartesian product $\mathcal{X}\times \mathcal{Y}$ is defined as $\{(x,y)|x\in\mathcal{X} \ \text{and} \ y\in\mathcal{Y}\}$ which contains a pair of elements from two sets.
A relation $r$ over sets $\mathcal{X}$ and $\mathcal{Y}$ can be defined to capture the relationships between $x$ and $y$.
Given a pair of images $(x,y)$ and 
 the corresponding
chronological ages $(\tau_x,\tau_y)$, we train a deep learning model to learn relations $r\in \mathcal{R}$ where $\mathcal{R}$ is the set of relationships without linear correlation, including ``cumulative relation" ($r_1=\tau_x+\tau_y$), ``relative relation" ($r_2 = \tau_x-\tau_y$), ``maximal relation" ($r_3=max(\tau_x,\tau_y)$), and ``minimal relation" ($r_4=min(\tau_x,\tau_y)$).
The cumulative relation $r_1=\tau_x+\tau_y$ aims to learn the sum of ages from the two subjects which can mitigate the additive noise.
If the model has errors on subjects $x$ and $y$ but the errors are in a different direction (positive on one and negative on the other), the cumulative relation $r_1=\tau_x+\tau_y$ will be close to the ground-truth.
Another potential application of the cumulative relation is the bias correction~\cite{de2020commentary}.
Based on the regression to the mean problem~\cite{barnett2005regression}, the brain ages of the young subjects are usually over-estimated and the brain ages of the old subjects are usually under-estimated. 
Recent studies are on correcting this bias~\cite{liang2019investigating,de2020commentary,beheshti2019bias}, but with controversies~\cite{butler2021pitfalls}.
With the pair of the young and old subjects, the cumulative relation we propose here may serve as a potential model for bias correction.
On the other hand, if the errors on two subjects are in the same direction, the relative relation $r_2=\tau_x-\tau_y$ may get us closer to the ground-truth. 
In addition, the relative relation can also be converted into the order relationship~\cite{lim2019order} or relative attributes~\cite{parikh2011relative}.
The maximal relation $r_3=\text{max}(\tau_x,\tau_y)$ and minimal relation $r_4=\text{min}(\tau_x,\tau_y)$ are useful for obtaining the upper and lower boundary of the estimated ages from the two subjects.

The advantages of the proposed deep relation learning are summarized as follows:
(1) it is an extension of the order learning. When only ``relative relation" $r_2=\tau_x-\tau_y$ is applied, it becomes the deep order learning~\cite{lim2019order};
(2) the values of the $x$ and $y$ can be directly estimated from the relations $r_i,i\in\{1,2,3,4\}$;
(3) mathematically, a relation is reflexivity if each element is related to itself.
The proposed four relations are reflexivity~\cite{wang2021pairwise} because the model can compute the relations between the pair of the same image $(x,x)$ and $r_1=2\tau_x$, $r_2=0$, $r_3=r_4=\tau_x$.
All the four relations can be used for prediction with an ensemble principle, similar to the multi-input multi-output (MIMO) method~\cite{havasi2020training};
(4) $y$ can be used as the reference instance with the known age $\tau_y$ to estimate $\tau_x$ based on ensemble learning, providing a robust prediction with different $y$.

We use the proposed deep relation learning for brain age estimation based on structural magnetic resonance imaging (MRI), which contains not only the anatomy of the brain, but also the brain 
age information.
The MRI-based brain age estimation is a typical regression task that aims at estimating the (artificial) biological brain age on brain MRIs using machine learning techniques~\cite{gaser201910,armanious2021age,cheng2021brain}.
The estimated ``brain age" is purely computed from brain MRIs and is a useful biomarker for brain health.
Mathematically, we use the $\nu$ and $\hat{\nu}$ to denote the chronological and estimated brain ages, respectively. 
$\hat{\nu}$ is estimated by machine learning model $\mathcal{M}$, given an input image $x$: $\hat{\nu}=\mathcal{M}(x)$.
The difference between the estimated and chronological brain ages $d=\hat{\nu}-\nu$ is usually named ``brain age gap, (BAG)"~\cite{cole2017predicting}.
Many studies have shown that the BAG is correlated to brain diseases or disorders, such as Alzheimer's Disease (AD)~\cite{brainwaa160}, Bipolar disorder~\cite{tonnesen2020brain}, Autism Spectrum Disorder (ASD)~\cite{tuncc2019deviation}, Psychopathology~\cite{cropley2021brain}, Major Depressive Disorder (MDD)~\cite{han2020brain}, Multiple sclerosis~\cite{hogestol2019cross}, Psychosis~\cite{chung2018use} and other common brain disorders~\cite{kaufmann2019common}.

Deep convolutional neural networks can extract task-discriminative features and learn the subtle patterns in the minimally pre-processed input MR images~\cite{abrol2021deep}.
Many deep learning brain MRI age estimation models have roots in models widely used in computer vision.
For example, the Age-Net~\cite{armanious2021age} which is a hybrid combination of inception v1~\cite{szegedy2015going} and SqueezeNet~\cite{iandola2016squeezenet}.
The DeepBrainNet~\cite{brainwaa160} which has been built based on the inception-resnet-v2~\cite{szegedy2017inception} framework, the two-stage-age-network (TSAN)~\cite{cheng2021brain} which is inspired by the DenseNet~\cite{huang2017densely}, the simple fully convolution network (SFCN) which is the lightweight version of the VGGNet~\cite{simonyan2014very} and the fusion with attention (FiA-Net)~\cite{he2021multi} which is the multi-channel fusion network based on the ResNet~\cite{he2016deep} and Hi-Net~\cite{zhou2020hi}.
All of these methods use deep learning models to estimate the brain age directly from input brain MRIs.
Our proposed method, however, is different from these models and aims to learn the different relations on a pair of input subjects. 
Our proposed method can also directly estimate the brain age from the pair of identical input images with an ensemble strategy. 

Fig.~\ref{fig:framework} shows the proposed hybrid neural network of deep relation learning for brain age estimation with a pair of input images $(x,y)$.
To learn different relations between two input subjects of brain MRIs, we first use a simple and efficient neural network (SFCN)~\cite{peng2019accurate} to extract the deep features.
The SFCN is a lightweight neural network and it can 
estimate brain ages from
3D MRIs with few parameters and low consumption of computational memories.
The Transformer with attention~\cite{vaswani2017attention} is exploited to learn different relations with a multi-task framework~\cite{liu2019multi}. 
A Transformer model in our proposed framework
uses the self-attention mechanism to learn the information from a pair of input images for relation learning.
Studies~\cite{yu2019multimodal,gabeur2020multi,prakash2021multi} have shown that Transformer can capture the inherent relations between different inputs.

The main contributions of this paper are summarized as follows:
\begin{itemize}
    \item we propose a deep relation learning framework for regression with four different relations not linearly related: accumulative, relative, maximal and minimal relations between a pair of input images.
    \item we evaluate the proposed deep relation learning for brain age estimation and there are different ways to estimate the brain age based on the pair of input images with the ensemble strategy.
    \item we propose a hybrid neural network with a convolutional neural network for feature extraction and a Transformer for relation learning.
\end{itemize}

The rest of the paper is organized as follows:
Section~\ref{sec:method} describes the proposed four different relations and the structure of the neural network for deep relation learning.
Section~\ref{sec:experi} presents the detailed experimental setting for brain age estimation using the proposed deep relation learning.
The results are provided in Section~\ref{sec:results} and the discussion and conclusion is given in Section~\ref{sec:conclusion}.

\section{Method}
\label{sec:method}

\subsection{Relations for regression}
As mentioned above, the Cartesian product $\mathcal{X}\times \mathcal{Y}$ is defined as $\{(x,y)|x\in\mathcal{X} \ \text{and} \ y\in\mathcal{Y}\}$ which contains a pair of elements from two sets $\mathcal{X}$ and $\mathcal{Y}$.
In this paper, the sets $\mathcal{X}=\mathcal{Y}$ are datasets containing brain MRIs and the defined relation is called a homogeneous relation~\cite{schmidt2012relations}.
For brain age estimation, the brain age range of the subjects is: $0\leq \tau_x,\tau_y\leq A$, where $A$ denotes the maximum age contained in the dataset. 
The typical relations can be defined as:
\begin{equation}
\label{eq:relations}
    \begin{array}{rcll}
    \text{Cumulative relation}: &  r_1 =&\tau_x + \tau_y  & r_1\in[0,2A]\\
    \text{Relative relation}: &    r_2 =& \tau_x - \tau_y & r_2\in[-A,A] \\
    \text{Maximal relation}: &    r_3 =& max(\tau_x,\tau_y) & r_3\in[0,A]\\
    \text{Minimal relation}: &    r_4 =& min(\tau_x,\tau_y) & r_4\in[0,A]\\
    \text{Amplifying relation}: & r_5 =& \tau_x*\tau_y & r_5\in[0,A^2]\\
    \text{Divided relation}: & r_6=&\tau_x/\tau_y & r_6\in[0,+\infty]\\
    \end{array}
\end{equation}
where $x$ and $y$ denote different input images and $\tau_x$ and $\tau_y$ denote the corresponding chronological ages of subjects $x$ and $y$, respectively.

In practice when using the neural network to estimate the relation and it should be bounded: $|r|\leq M$ where $M$ is a real number.
Thus, we only use the first four relations $r_i,i\in\{1,2,3,4\}$ and do not consider the ``amplifying relation $r_5$" and the ``divided relation $r_6$" since their boundaries are very large, toward [0,10,000] or even [0,+$\infty$], making it hard to train the neural network on a lifespan dataset with brain age of 0-100 years.
We train a neural network to learn the four relations $r_i,i\in\{1,2,3,4\}$ and the brain age estimation can be obtained based on the estimated $r_i,i\in\{1,2,3,4\}$ from the trained neural network.

\subsection{Deep neural network}
\label{sec:cnn}

The framework of the proposed deep relation learning is shown in Fig.~\ref{fig:framework}, which contains two parts: a convolutional neural network (CNN) backbone for deep feature extraction from the pair of input images and a Transformer for the fusion of the extracted deep features to learn the four relations.

\subsubsection{CNN backbone to extract deep features}
For the CNN backbone, we use a structure similar to the Simple Fully Convolutional Neural Network (SFCN)~\cite{peng2019accurate}.
The network contains 6 blocks and each block on the first 5 blocks contains a convolutional layer with a kernel size of $3\times 3\times 3$, a batch normalization layer~\cite{ioffe2015batch}, a ReLU activation layer~\cite{krizhevsky2012imagenet} and a max-pooling layer with a kernel size of $2\times 2\times 2$ and stride $2$.
The last block contains a convolutional layer with a kernel size of $1\times 1\times 1$, a batch normalization and a ReLU layer. 
We also set the channel numbers of each convolutional layers to $[32,64,128,256,256,64]$, as in~\cite{peng2019accurate}.

In practice we have found that applying a max-pooling layer with the kernel size of $2\times 2\times 2$ and stride $2$ at the beginning of the neural network or directly on the input images (see Fig.~\ref{fig:framework}) can 
reduce the errors in brain age estimation.
It can also reduce the image size at the beginning and thus reduce the computational complexity and memory cost. 
Applying the max-pooling directly on input images can help reduce the redundant information in the input images.
We name the neural network with the max-pooling at the first layer ``mSFCN" in this paper.
Using the CNN, the 3D input image $x$ can be converted into a tensor $\textbf{T}^{d\times h\times w\times c}_x$ where $d=64$ is the feature dimension and $h,w,c$ are the height, the width, and the number of channels in each input image, respectively.

\subsubsection{Transformer to learn relations}

Given 4D feature tensors $\textbf{T}_x$ and $\textbf{T}_y$ extracted from the pair of input images $(x,y)$, we exploit a ``patch" operation to convert the feature tensors into a sequence of feature vectors where each feature vector represents the deep feature from a patch receptive field of the input image.
These two 4D feature tensors are reshaped into 2D sequences of tokens:
$\textbf{t}^{d\times L}_x$ and $\textbf{t}^{d\times L}_y$ (where $L=h\times w\times c$).
Each token is a 1D feature vector with the size of $L$ and there are $L=h\times w\times c$ tokens in total extracted on each 4D feature tensor.
In practice, the size of the input image is $2\times 80\times 130 \times 170$ where $2$ represents the 2 channels of input images (intensity image and RAVENS map, see details on Section~\ref{sec:experi}.A) and $80,130,170$ are the size of the three dimensions of the brain MRIs.
The size of the tensors $\textbf{T}_x$ and $\textbf{T}_y$ extracted from the mSFCN is $(d=) 64 \times (h=) 4 \times (w=) 5 \times (c=) 2$ after five max-pooling layers (with kernel size of 2 and stride of 2) on the mSFCN backbone.
Thus, the size of the corresponding token sequences $\textbf{t}_x$ and $\textbf{t}_y$ is $64\times 40$.
These two sequences of tokens are concatenated into a sequence of $2L$ tokens~\cite{dosovitskiy2020image}: $\textbf{t}^{d\times 2L}=[\textbf{t}^{d\times L}_x,\textbf{t}^{d\times L}_y]$.
The combined tokens from the two input images are concatenated into a sequence of tokens with a size of $64\times 80$ and fed into a standard Transformer~\cite{vaswani2017attention} which contains several encoder blocks.
Each encoder block contains two parts: an attention part and a multi-layer perceptron (MLP) part.
The attention part consists of Layernorm and multi-headed self-attention (the head size is 8) layers:
\begin{equation}
    \textbf{t}^{d\times 2L}=\text{Self-Attention}(\textbf{t}^{d\times 2L})
\end{equation}
The key idea of self-attention is to fuse the tokens with the attention mechanism.
It first transforms the sequence of tokens $\textbf{t}^{d\times 2L}$ into query $Q^{d\times 2L}$, key $K^{d\times 2L}$ and value $V^{d\times 2L}$ by a linear transformation:
\begin{equation}
    \text{Attention}(Q,K,V)=\text{Softmax}(\frac{QK^T}{\sqrt{d}})V
\end{equation}

The second part is multi-layer perceptron (MLP) layer. It is also named Feed-Forward Networks (FFN)~\cite{vaswani2017attention} which contains two fully connected layers with a ReLU activation in between:
\begin{equation}
    \textbf{t}^{d\times 2L}=\text{FFN}(\textbf{t}^{d\times 2L})
\end{equation}

Finally, the outputs of the relation estimation are computed as:
\begin{equation}
    \hat{r}_i=\text{class\_head}(\textbf{t}^{d,i})
\end{equation}
where $\textbf{t}^{d,i}$ is the $i$th token sampled from the sequence of the tokens $\textbf{t}^{d\times 2L}$ and $\text{class\_head}$ is a fully-connected layer to learn the relation $r_i,i\in\{1,2,3,4\}$.

\subsubsection{Neural network training}

We use the mean absolute error (MAE) as the loss function to train the neural network for relation learning:
\begin{equation}
\label{eq:mae}
    \mathcal{L}_i = \frac{1}{n}\sum_{j=1}^n|\hat{r}_i-r_i|
\end{equation}
where $\hat{r}_i$ is the estimated relation from the network and $r_i$ is the ground-truth for the $i$th relation ($1\leq i\leq 4$).
$n$ is the number of samples involved in the computation (i.e. the number of training samples on each batch).
The mean absolute error loss is widely used to train the neural network for brain age estimation~\cite{jonsson2019brain,feng2020estimating,armanious2021age}.
The neural network is trained with a total loss computed by:
\begin{equation}
    \mathcal{L}=\sum_{i=1}^K \mathcal{L}_i
\end{equation}
where $K$ is the number of relations estimated by the neural network.
For joint relation learning, the neural network is used to estimate the four relations and $K=4$.
For pair relation learning, the neural network is trained to estimate a pair of relations ($(r_1,r_2)$ and $(r_3,r_4)$) and $K=2$.
For single relation learning, the neural network is only used to estimate one relation and $K=1$.
The detailed information of the joint, pair, and single relation learning can be found in Section~\ref{sec:relationest}.

\section{Experiments}
\label{sec:experi}

We will first introduce our data (\ref{sec:dataset}) and network training (\ref{sec:netrain}).
Two major experiment settings follow: optimizing various components in the framework to maximize the accuracy of relation learning (\ref{sec:opt1}), and evaluating the accuracy of the proposed relation-based brain age estimation (\ref{sec:acp02}).

\subsection{Data sets}
\label{sec:dataset}

\begin{table}[!t]
    \centering
    \caption{Demographics of datasets used in this paper, sorted by median age (years).}
    \label{tab:dataset}
    \begin{tabular}{l|ccc}
    \toprule
         Dataset &  N$_{\text{samples}}$ & Age range [Median] & Male/Female \\
    \midrule
         MGHBCH~\cite{he2020brain} & 428 & 0-6 [1.7] & 226/202 \\
         NIH-PD~\cite{evans2006nih} & 1,211 & 0-22.3 [9.8] & 585/626 \\
         ABIDE-I~\cite{di2014autism}  & 567 & 6.47-56.2 [14.8]  & 469/98  \\
         BGSP~\cite{holmes2015brain} & 1,570 & 19-53 [21] & 665/905  \\
         BeijingEN~\footnote{} & 180 & 17-28 [21] & 73/107 \\
         IXI~\footnote{} & 556 & 20.0-86.3 [48.6] &  247/309 \\
         DLBS~\cite{park2012neural} & 315 & 20-89 [54] & 117/198 \\
         OASIS-3~\cite{lamontagne2018oasis} & 1,222 & 42-97 [69] & 750/472 \\
        \cmidrule{1-4}
        Total & 6,049 & 0-97 [22.8] & 3,132/2,917 \\
    \bottomrule
    \multicolumn{4}{l}{1, \url{http://fcon_1000.projects.nitrc.org/indi/retro/BeijingEnhanced.html}}\\
    \multicolumn{4}{l}{2, \url{https://brain-development.org/ixi-dataset/}} \\
    \end{tabular}
\end{table}

\begin{figure}[!t]
    \centering
    \includegraphics[width=0.4\textwidth]{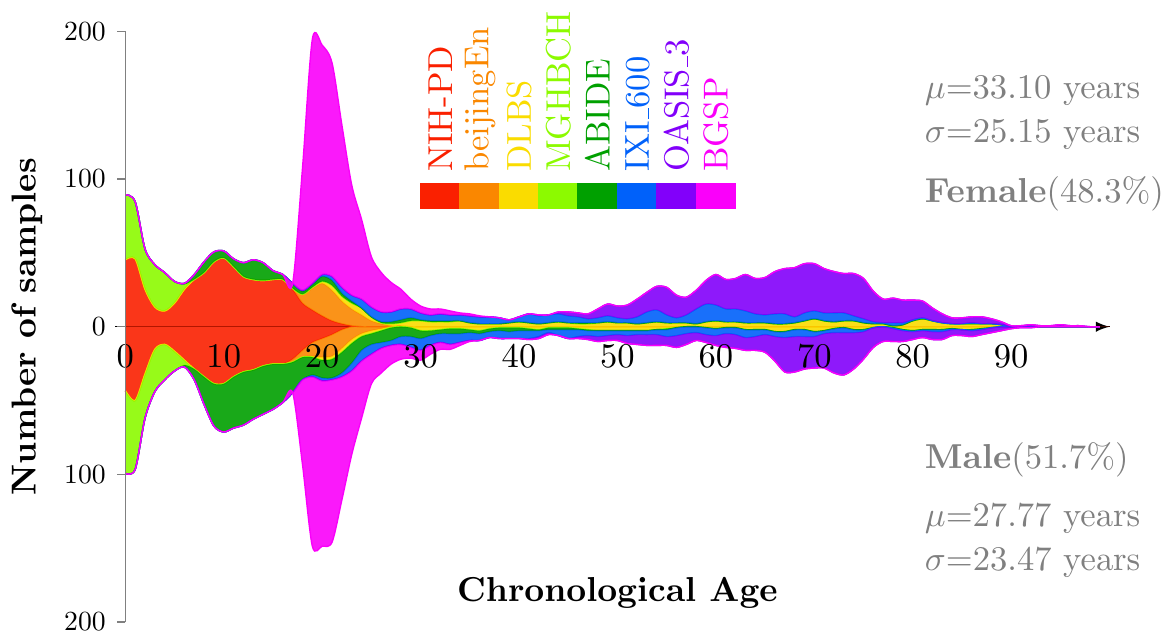}
    \caption{Age distribution of the datasets used in this paper, covering the age from 0 to 97 years, with a mean of $\mu=30.58$ years (median 22.8 years) and a standard deviation of $\sigma=24.52$ years.}
    \label{fig:datadistribution}
\end{figure}

In this paper, we use a lifespan dataset which is the same as used in our previous work~\cite{he2021multi}.
A lifespan dataset is recently used in studies~\cite{becker2018gaussian,pomponio2020harmonization,brainwaa160}
and the deep learning model trained on a lifespan dataset can be applied/transferred to any age group
without introducing artificial boundaries on the predicted ages.
This is especially important for quantifying premature aging or development delays in diseased cohorts, as MRI-manifested brain ages can range from 0-100 years even though the patient's actual ages are in a narrowly bounded range.
The summary of the dataset is shown in Table~\ref{tab:dataset}.
The merged dataset consists of brain MRI scans from 8 datasets with 6,049 samples (0-97 years of age).
Fig.\ref{fig:datadistribution} shows the age distribution of the dataset.
Only healthy brains with T1-weighted MRIs are collected in each dataset.

Similar to~\cite{he2021multi}, we perform a minimum pre-processing and harmonization of the T1w images with the following steps:
(1) N4 bias correction~\cite{tustison2010n4itk}; (2) field of view normalization~\cite{ou2018field}; 
(3) Multi-Atlas Skull Stripping (MASS)~\cite{doshi2013multi,ou2015brain};
(4) non-rigidly registered to the SRI24 atlas \cite{rohlfing2010sri24} by the Deformable Registration via Attribute-Matching and Mutual-Saliency weighting (DRAMMS) algorithm~\cite{ou2011dramms};
(5) splitting the registered image into two channels: intensity image containing the \textit{contrast} information and RAVENS map~\cite{davatzikos2001voxel} containing the \textit{morphological} information.
The final size of the MRIs after pre-processing is $80\times 130\times 170$ per channel without the black background voxels on the boundary.
We concatenate the intensity image and RAVENS map as the input of the 3D neural network so the image size of each subject is $2\times 80\times 130\times 170$.
Our previous study showed that explicitly splitting the T1w image into two channels led to more accurate age estimation than each channel alone~\cite{he2021multi}.

\subsection{Network training}
\label{sec:netrain}

\begin{figure}[!t]
    \centering
    \includegraphics[width=0.4\textwidth]{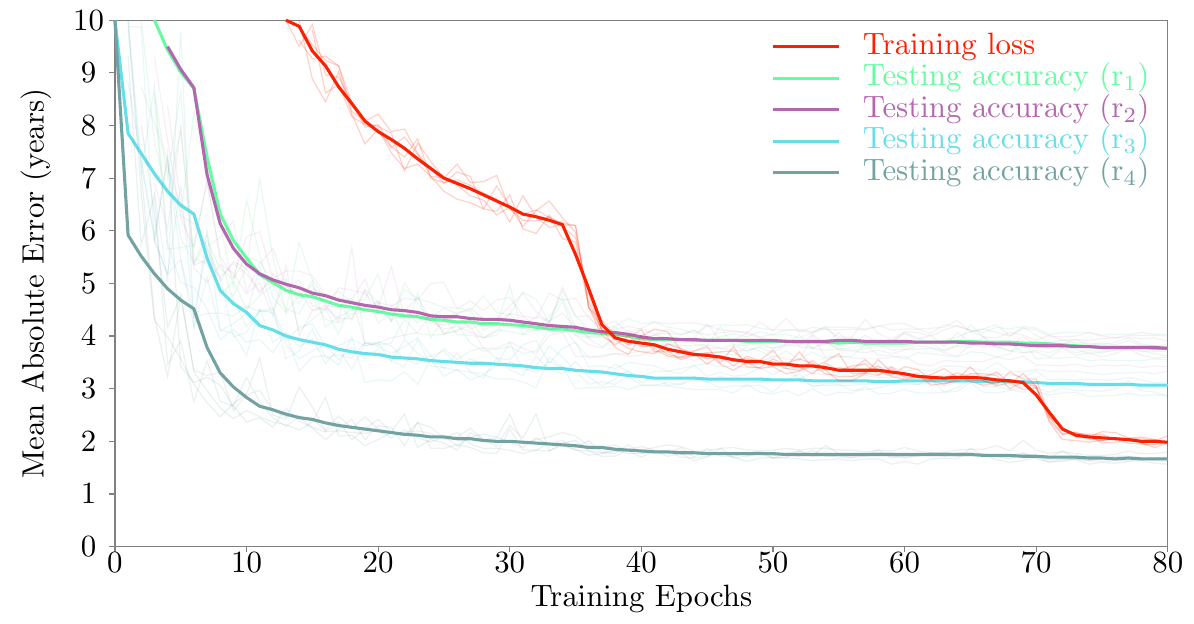}
    \caption{The training loss and testing accuracies (in terms of MAE, the lower MAE, the better accuracy) of different relations. The bold curves are the average MAEs over the five-fold cross-validation.}
    \label{fig:trainloss}
\end{figure}

The network is trained by the Adam optimizer built in PyTorch platform, with an initial learning rate of 0.0001, reducing to half at every 35 epochs in the total 80 training epochs.
The batch size is set to 20 due to the limitation of the GPU memory.
The training of the neural network takes around 24 hours on a single NVIDIA RTX 6000 GPU with 24G memory.
We divide the training images into 100 groups according to their ages.
To make a balanced distribution of the input pair $(x,y)$ during training, on each iteration, we first randomly select an age group and then randomly select an image from this group to collect the batch of the training samples.
Fig.~\ref{fig:trainloss} shows the training loss and testing accuracies of the four relations. 
The MAEs of the four relations decrease fast in the first 35 epochs.
After that, the neural network starts to converge and the accuracies of the four relations increase slowly and are stable with 80 training epochs.

\subsection{Optimization of the accuracy for relationship estimation}
\label{sec:opt1}
\subsubsection{Accuracy metric in cross validation for relationship estimation}
\label{sec:C1}

We use the cross-validation strategy~\cite{schaffer1993selecting,he2021multi} to evaluate the accuracy of the relation estimation.
The merged dataset is randomly split into 5 folds of approximately equal sample sizes without overlapping. 
Each time, one fold of the samples is used for evaluation and the rest four folds are used for training.
It repeats five times and each sample is left out for testing once and only once.

Our relations are computed based on a pair of input images and the evaluation of the relation estimation is performed on a set of test pairs which contains $N$ pairs of input images.
In other words, $N$ is the total number of testing pairs involved in the evaluation.
We use three accuracy metrics computed on the test pairs: mean absolute error (MAE), cumulative score (CS) and Pearson correlation coefficient.
The MAE is computed as the mean absolute errors between the estimation and ground-truth of the relations on test pairs.
Note that the MAE is computed on the test samples which is different from the MAE loss defined in Eq.~\ref{eq:mae} computed on the training samples.
The CS is computed by: $CS(\alpha)=N_{|e|\leq\alpha}/N\times 100\%$, 
where $N_{|e|\leq\alpha}$ is the number of test pairs whose absolute error $|e|$ is no higher than a given threshold $\alpha$.
Following previous works~\cite{he2021multi,he2021global}, we set  $\alpha=5$ (years) in experiments.
The Pearson correlation is computed between the ground-truth $r_i$ and the estimated $\hat{r}_i$ ($i\in\{1,2,3,4\}$) on the whole test set.

\subsubsection{Eight variations of underlying deep learning models}
There are two main parts of the neural network for relation learning given a pair of input $(x,y)$: feature extraction and relation regression (as shown in Fig.~\ref{fig:framework}).
For the feature extraction part,  we compare the accuracies of the   SFCN~\cite{peng2019accurate} and the mSFCN which applies a Max-Pooling layer (with the kernel size of $2\times2\times2$, stride size of $2$) on the input image.
The backbones $\mathcal{F}_1$ and $\mathcal{F}_2$ can be shared (similar to the Siamese network~\cite{lim2019order}: $\mathcal{F}_1=\mathcal{F}_2$) or independent (the same network with different parameters~\cite{havasi2020training}: $\mathcal{F}_1\neq \mathcal{F}_2$).
For the relation regression part, we consider the traditional CNN-based method~\cite{lim2019order} as the baseline: deep features from the two pairs are concatenated and input to three fully-connected (FC) layers with the size of: 64, 64, 4-channel vectors sequentially.
The 4-channel vectors represent the 4 output relations.
We also use the Transformer to compute the four relations and we set the number of transformer blocks to 2, with the same number of FC layers of the CNN-based method for a fair comparison.

Table~\ref{tab:modelvarious} summarizes the 8 model variations compared in this paper and the main difference is which backbone (SFCN or mSFCN, and whether it is shared or independent) is used for feature extraction and which model (FCs or Transformer) is used for relation regression (Fig.~\ref{fig:framework}).

\begin{table}[!t]
    \centering
        \caption{Model variations compared  with different backbones and relation regressions. (FCs represents the three Fully-Connected layers for relation regression)}
    \label{tab:modelvarious}
    \resizebox{0.5\textwidth}{!}{
    \begin{tabular}{l|l}
    \toprule
       Model Name  &  Configuration\\
    \midrule
    SFCN$_s$+FCs & Shared SFCN backbone ($\mathcal{F}_1=\mathcal{F}_2$) with FCs for relation regression \\
    SFCN$_i$+FCs & Independent SFCN backbone ($\mathcal{F}_1\neq\mathcal{F}_2$) with FCs for relation regression \\
    mSFCN$_s$+FCs & Shared mSFCN backbone ($\mathcal{F}_1=\mathcal{F}_2$) with FCs for relation regression \\
    mSFCN$_i$+FCs & Independent mSFCN backbone ($\mathcal{F}_1\neq\mathcal{F}_2$) with FCs for relation regression \\
    SFCN$_s$+Transformer & Shared SFCN backbone ($\mathcal{F}_1=\mathcal{F}_2$) with Transformer for relation regression \\
    SFCN$_i$+Transformer & Independent SFCN backbone ($\mathcal{F}_1\neq\mathcal{F}_2$) with Transformer for relation regression \\
    mSFCN$_s$+Transformer & Shared mSFCN backbone ($\mathcal{F}_1=\mathcal{F}_2$) with Transformer for relation regression \\
    mSFCN$_i$+Transformer & Independent mSFCN backbone ($\mathcal{F}_1\neq\mathcal{F}_2$) with Transformer for relation \\
    \bottomrule
    \end{tabular}}
\end{table}

\subsubsection{Using 1, 2 or 4 deep learning models for estimating 4 relations}
\label{sec:relationest}

The outputs of the proposed neural network are the four relations of the pair of inputs $(x,y)$. 
Three different configurations are considered: joint relation learning, pair relation learning and single relation learning.
For the joint relation learning, we train 1 neural network to learn the four relations with the multi-task framework~\cite{ruder2017overview}.
For the pair relation learning, we train 2 neural networks to learn the pair of relation ($r_1$, $r_2$) and ($r_3$, $r_4$), respectively.
For the single relation learning, we train 4 neural networks and each neural network learns one relation. 
On the test set, the pairs of the input $(x,y)$ are randomly sampled for the testing set and the results from the 5-fold cross-validation are averaged.

\subsection{Accuracy evaluation for brain age estimation}
\label{sec:acp02}

\subsubsection{Accuracy metric in cross-validation for brain age estimation}
\label{sec:D1}
Similar to Section~\ref{sec:opt1}, we also use 5-fold cross-validation strategy to evaluate the accuracy of brain age estimation based on estimated relations in different configurations given a pair of input images.

For evaluation of brain age estimation, we also use the three accuracy metrics: MAE, CS and Pearson correlation coefficient.
Specifically, the MAE for brain age estimation is computed by: $\text{MAE}=\frac{1}{M}\sum_{j=1}^M|\hat{\tau}_x-\tau_x|$ where $\hat{\tau}_x$ is the estimated brain age of the test scan $x$ and $\tau_x$ is the corresponding chronological age.
$M$ is the total number of the testing images.
The CS for brain age estimation is computed by: $CS(\alpha)=M_{|e|\leq\alpha}/M\times 100\%$
which indicates the accuracy of brain age estimation with the absolute error $|e|$ no higher than a given threshold $\alpha$.
We also set the $\alpha=5$ (years) for brain age estimation in experiments.
The Pearson correlation is computed between the ground-truth $\tau_x$ and the estimated $\hat{\tau}_x$ on the whole test images (M=6,049 in our experiment since each subject has been left out once and only once in the 5-fold cross-validation).

To measure the significance of the improvement of the proposed method compared to other state-of-the-art models, we use paired $t$-test (two-side) to compute $p$-value between the absolute errors obtained by the proposed method and absolute errors obtained by other state-of-the-art models in comparison over all the 6,049 test images for brain age estimation.
$p\le 0.05$ indicates that there is a significant difference between the MAEs of the two different models in comparison.

\subsubsection{Accuracy of relation-based brain age estimation when $x\neq y$, and $x$ and $y$ both from the testing set}
\label{sec:D2}
Based on the estimated relations $\hat{r}_i,i\in\{1,2,3,4\}$ of the input pair $(x,y)$, the estimated brain age $\hat{\tau}_x$ and $\hat{\tau}_y$ of both $x$ and $y$ can be computed simultaneously based on the following equations:

\begin{equation}
\label{eq:xny}
    \begin{array}{rl}
        \hat{\tau}_x =&(\hat{r}_1+\hat{r}_2)/2 \\
        \hat{\tau}_y =&(\hat{r}_1-\hat{r}_2)/2 \\
        \hat{\tau}_x = &
        \begin{cases}
        \hat{r}_3 &\text{if} \ \ \ \ \  \hat{r}_2>0\\
        \hat{r}_4 &\text{Otherwise}\\
        \end{cases} \\
        \hat{\tau}_y = &
        \begin{cases}
        \hat{r}_4 &\text{if} \ \ \ \ \ \hat{r}_2<0\\
        \hat{r}_3 &\text{Otherwise}\\
        \end{cases}
    \end{array}
\end{equation}
There are two estimations of each pair $x$ and $y$ and 
we estimate the age of image x by averaging the two $\hat{\tau}_x$ above, and do the same for estimating the age for image y.

In this section, we group the testing images into pairs and feed them into the trained neural network for relation estimation. 
The estimated brain ages of both $x$ and $y$ are computed according to Eq.(\ref{eq:xny}). 
The average of the MAE, CS($\alpha$=5 years) and Pearson correlation from models trained for 5-fold cross-validation are reported.

\subsubsection{Accuracy of relation-based brain age estimation on $x$ with reference $y$ when $x\neq y$, $x$ from the testing set and $y$ from the training set.}
In this section, we estimate the brain age $\hat{\tau}_x$ of the testing input $x$ from the testing set based on the reference $y$ which is sampled from the training set with known brain age $\tau_y$.
The estimation of $\tau_y$ is close to the ground-truth because $y$ is sampled from the training set.
Thus, the error of the estimation $\hat{\tau}_x$ is only from the test sample $x$.
We sample no more than 2 brain MRIs from each age on the training set and there are roughly 186 reference samples $y$ used in experiments.
Based on the definition the four relations as defined on Eq.~(\ref{eq:relations}), for each input testing image $x$, there are four different ways to estimate the brain age $\hat{\tau_x}$ of $x$ based on the known age $\tau_y$ of the reference $y$:
\begin{equation}
\label{eq:e10}
    \begin{array}{rl}
        \hat{\tau}_x =&\hat{r}_1-\tau_y  \\
        \hat{\tau}_x =&\hat{r}_2+\tau_y \\
        \hat{\tau}_x =&(\hat{r}_1+\hat{r}_2)/2 \\
        \hat{\tau}_x =&\hat{r}_3+\hat{r}_4-\tau_y \\
    \end{array}
\end{equation}
where $\hat{r}_i,i\in\{1,2,3,4\}$ are the estimation of the four relations from the trained neural network.
We also compare the proposed methods with the baseline model proposed in~\cite{lim2019order}, which only considers the ``relative relation $r_2$" and converts it into a binary relation given a threshold $t$~\cite{lim2019order}: $\tau_x>\tau_y$ if $r_2>t$, $\tau_x\approx \tau_y$ if $|r_2|\leq t$ and $\tau_x<\tau_y$ if $r_2<-t$.
Given the binarized order relationship, the estimation of $\hat{\tau_x}$ is obtained by the maximum consistency (MC) rule~\cite{lim2019order} given $N$ different reference $y$ with the chronological age $\tau_{y_i}$:
\begin{equation}
    \hat{\tau}_x=\arg \max_{\tau_{x'}} \sum_{i=0}^{N-1} \phi(\tau_x,\tau_{y_i},\tau_{x'})
\end{equation}
where $\phi(\tau_x,\tau_{y_i},\tau_{x'})$ is the consistency function defined by:
\begin{equation}
\begin{array}{rl}
    c_1 =&[r_2>t][\tau_{x'}-\tau_{y_i}>t]  \\
    c_2 =&[|r_2\leq t][|\tau_{x'}-\tau_{y_i}|\leq t] \\
    c_3 =& [r_2<-t][\tau_{x'}-\tau_{y_i}<-t] \\
    \phi(\tau_x,\tau_{y_i},\tau_{x'})=& c_1+c_2+c_3  \\
\end{array}
\end{equation}
where $[\cdot]$ is the indicator function and $\phi(\tau_x,\tau_{y_i},\tau_{x'})$ returns either 0 (inconsistent) or 1 (consistent).
More computation and explanation details of the MC rule can be found in~\cite{lim2019order}.
The complexity of the MC rule is $\mathcal{O}(MN)$ where $M$ is the number of references  and $N$ is the total age bins ($N=97$ in this paper, covering the age from 0-97 years of age).

\subsubsection{Accuracy of relation-based brain age estimation when $x=y$, and both from the testing set}

The pair of inputs $(x,y)$ can be the same testing image when we set the input $y=x$.
Ideally, when $y=x$, the ``relative relation" $r_2=0$ and $r_3=r_4=x$.
However, due to the possible regression errors from the neural network, the estimated relations $\hat{r}_2$ might not be zeros and $\hat{r}_3\neq\hat{r}_4$ may occur.
Thus, all estimations of the four relations $\hat{r}_i,i\in\{1,2,3,4\}$ can be used to estimate the brain age $\hat{x}$ based on the following calculations:
\begin{equation}
\label{eq:selfcom}
    \begin{array}{rl}
        \hat{\tau_x} =& \hat{r}_1/2  \\
        \hat{\tau_x} =& (\hat{r}_1+\hat{r}_2)/2  \\
        \hat{\tau_x} =& (\hat{r}_1-\hat{r}_2)/2  \\
        \hat{\tau_x} =& \hat{r}_3  \\
        \hat{\tau_x} =& \hat{r}_4  \\
        \hat{\tau_x} =& (\hat{r}_3+\hat{r}_4)/2  \\
    \end{array}
\end{equation}

\subsubsection{Accuracy comparison with state-of-the-art brain age estimation algorithms}

Eight other deep learning based methods for brain age estimation are compared in this section, including the Hi-Net~\cite{zhou2020hi}, FiA-Net~\cite{he2021multi}, GL-Transformer~\cite{he2021global}, 3D CNN~\cite{cole2017predicting}, SFCN~\cite{peng2019accurate}, DeepBrainNet~\cite{brainwaa160}.
The Hi-Net~\cite{zhou2020hi} and FiA-Net~\cite{he2021multi} fuse the multi-channel input MRI images (intensity and RAVENS) in a layer-level fusion.
The GL-Transformer~\cite{he2021global} applies a global-local transformer for exploiting the global-context information from the whole image and the local fine-grained information from the local patches on 2D input images.
The 3D CNN~\cite{cole2017predicting} uses 5 convolutional layers with kernel size of $3\times 3\times 3$, followed by ReLU and max-pooling layers.
The number of channels at the first layer is eight, and is doubled after each max-pooling layer.
We use the global average pooling layer and a fully-connected layer for the brain age estimation.
The SFCN\cite{peng2019accurate} is also used as the backbone in our method, as described in Section~\ref{sec:cnn}.
The DeepBrainNet~\cite{brainwaa160} is based on the Inception-Res-V2~\cite{szegedy2017inception} model, which works on 2D slices for brain age estimation.
We get the results of Hi-Net~\cite{zhou2020hi} and FiA-Net~\cite{he2021multi} from the original research papers since we have similar experimental configurations and datasets.
We also compare the accuracies of models with the relation learning (mSFCN+Transformer+Relation) and without relation learning (mSFCN+Transformer) where the model is directly trained to estimate the brain age.
For the GL-Transformer~\cite{he2021global}, 3D CNN~\cite{cole2017predicting}, SFCN~\cite{peng2019accurate}, DeepBrainNet~\cite{brainwaa160} and mSFCN+Transformer, we train them from scratch with the same training configuration as our neural network for a fair comparison.

\section{Results}
\label{sec:results}
This section has two major parts:
Section~\ref{sec:ac01} reports the accuracies for relation learning, which are results from experiment setting in~\ref{sec:opt1}; and Section~\ref{sec:ac02} presents the accuracies for relation-based brain age estimation, which are results from experiment setting introduced in Section~\ref{sec:acp02}.

\subsection{Accuracy of relation estimation}
\label{sec:ac01}

This section presents the accuracies of relation estimation which are measured by the three metrics of MAE, CS($\alpha$=5) and Pearson correlation described in Section~\ref{sec:C1} on the test samples.
Higher accuracy means lower MAE and higher scores of CS and Pearson correlation for relation estimation in this section.

\subsubsection{Effects of underlying CNN models}

\newcommand{\CC}{\cellcolor{gray!10}}
\begin{table*}[!t]
    \centering
    \caption{The accuracies of relation estimation $\hat{r}_i$ with pair samples from the test set.}
    \label{tab:relationres}
    \resizebox{\textwidth}{!}{
    \begin{tabular}{ll|cc|cc|cc|cc}
    \toprule
    \multicolumn{2}{c|}{\multirow{4}{*}{Different strategies}} & \multicolumn{4}{c|}{Relation regression based on FCs~\cite{lim2019order}} & \multicolumn{4}{c}{Relation regression based on Transformer}\\
    \cmidrule{3-10}
    & & \multicolumn{2}{c|}{CNN Backbone: SFCN~\cite{peng2019accurate}} & \multicolumn{2}{c|}{CNN Backbone: mSFCN} & \multicolumn{2}{c|}{CNN Backbone: SFCN~\cite{peng2019accurate}} & \multicolumn{2}{c}{CNN Backbone: mSFCN}  \\
    \cmidrule{3-10}
    & & Shared  & Independent  & Shared & Independent  & Shared  & Independent & Shared  & Independent  \\
    & & $\mathcal{F}_1=\mathcal{F}_2$ &  $\mathcal{F}_1\neq\mathcal{F}_2$ &  $\mathcal{F}_1=\mathcal{F}_2$ &  $\mathcal{F}_1\neq\mathcal{F}_2$ &  $\mathcal{F}_1=\mathcal{F}_2$ &  $\mathcal{F}_1\neq\mathcal{F}_2$&  $\mathcal{F}_1=\mathcal{F}_2$ &  $\mathcal{F}_1\neq\mathcal{F}_2$ \\
\midrule
\multirow{3}{*}{$\hat{r}_1$} & MAE & 4.88$\pm$0.45 & 4.69$\pm$0.30 & 5.57$\pm$0.74 & 5.08$\pm$0.51 & 4.13$\pm$0.24 & 4.74$\pm$0.28 & 3.71$\pm$0.18 & 3.88$\pm$0.14 \\
& CS($\alpha=5$)& 62.13\%$\pm$5.35\% & 63.15\%$\pm$3.75\% & 58.66\%$\pm$6.43\% & 61.89\%$\pm$5.41\% & 70.74\%$\pm$2.43\% & 64.60\%$\pm$3.38\% & 75.04\%$\pm$2.01\% & 73.18\%$\pm$1.63\% \\
& Pearson& 0.9851$\pm$0.0012 & 0.9844$\pm$0.0016 & 0.9871$\pm$0.0022 & 0.9845$\pm$0.0028 & 0.9862$\pm$0.0018 & 0.9822$\pm$0.0019 & 0.9878$\pm$0.0016 & 0.9866$\pm$0.0014\\
\midrule
\multirow{3}{*}{$\hat{r}_2$} & MAE & 5.40$\pm$0.53 & 5.69$\pm$0.29 & 5.65$\pm$0.39 & 5.70$\pm$0.47 & 5.63$\pm$0.12 & 5.01$\pm$0.44 & 3.80$\pm$0.24 & 3.88$\pm$0.14\\
& CS($\alpha=5$)& 59.22\%$\pm$4.81\% & 53.96\%$\pm$2.18\% & 57.51\%$\pm$3.69\% & 57.41\%$\pm$4.66\% & 57.96\%$\pm$1.21\% & 62.46\%$\pm$4.80\% & 74.17\%$\pm$2.06\% & 74.02\%$\pm$1.76\%\\
& Pearson& 0.9844$\pm$0.0011 & 0.9832$\pm$0.0018 & 0.9861$\pm$0.0016 & 0.9846$\pm$0.0036 & 0.9742$\pm$0.0020 & 0.9804$\pm$0.0028 & 0.9880$\pm$0.0018 & 0.9868$\pm$0.0017\\
\midrule
\multirow{3}{*}{$\hat{r}_3$} & MAE & 4.81$\pm$0.54 & 5.17$\pm$0.39 & 5.80$\pm$0.68 & 5.10$\pm$0.60 & 3.56$\pm$0.24 & 3.84$\pm$0.14 & 3.03$\pm$0.16 & 3.15$\pm$0.10 \\
& CS($\alpha=5$)& 61.88\%$\pm$4.41\% & 58.21\%$\pm$4.55\% & 54.07\%$\pm$5.30\% & 60.40\%$\pm$5.55\% & 76.36\%$\pm$1.55\% & 74.82\%$\pm$2.46\% & 81.62\%$\pm$1.44\% & 80.79\%$\pm$0.93\%\\
& Pearson& 0.9778$\pm$0.0023 & 0.9723$\pm$0.0035 & 0.9798$\pm$0.0032 & 0.9779$\pm$0.0026 & 0.9783$\pm$0.0043 & 0.9749$\pm$0.0020 & 0.9838$\pm$0.0027 & 0.9823$\pm$0.0015\\
\midrule
\multirow{3}{*}{$\hat{r}_4$} & MAE & 3.81$\pm$0.23 & 4.36$\pm$0.33 & 3.10$\pm$0.20 & 3.75$\pm$0.12 & 1.82$\pm$0.04 & 2.24$\pm$0.12 & 1.69$\pm$0.11 & 1.84$\pm$0.13 \\
& CS($\alpha=5$)& 72.83\%$\pm$3.03\% & 67.01\%$\pm$3.14\% & 82.18\%$\pm$1.28\% & 75.46\%$\pm$1.82\% & 93.09\%$\pm$0.50\% & 89.42\%$\pm$1.96\% & 94.19\%$\pm$0.36\% & 93.07\%$\pm$1.06\%\\
& Pearson& 0.9502$\pm$0.0050 & 0.9343$\pm$0.0090 & 0.9654$\pm$0.0023 & 0.9508$\pm$0.0029 & 0.9803$\pm$0.0028 & 0.9744$\pm$0.0036 & 0.9825$\pm$0.0044 & 0.9777$\pm$0.0066\\
\bottomrule
\end{tabular}}
\end{table*}

Table~\ref{tab:relationres} shows the accuracies of 8 model variations when the pair of input images are sampled from the testing set.
Several observations can be obtained:
(1) the MAEs of the ``maximal relation ($\hat{r}_3$)" and ``minimal relation ($\hat{r}_4$)" are lower than the MAEs of ``cumulative relation ($\hat{r}_1$)" and ``relative relation ($\hat{r}_2$)", indicating that the neural network is more powerful at capturing the non-linear relations than the linear relations of the pair of input images.
The ``minimal relations ($\hat{r}_4$)" has the lowest MAEs among the four relation estimations.
(2) Using Transformer for the relation regression provides better accuracies than using FCs for relation regression on different configurations.
(3) the mSFCN$_i$ provides the best accuracies than other model variations for the four relations regression.
(4) when using the Transformer for relation regression, the shared backbone $\mathcal{F}_1=\mathcal{F}_2$ provides better results than the independent backbone $\mathcal{F}_1\neq\mathcal{F}_2$.

\begin{figure}[!t]
    \centering
    \includegraphics[width=0.5\textwidth]{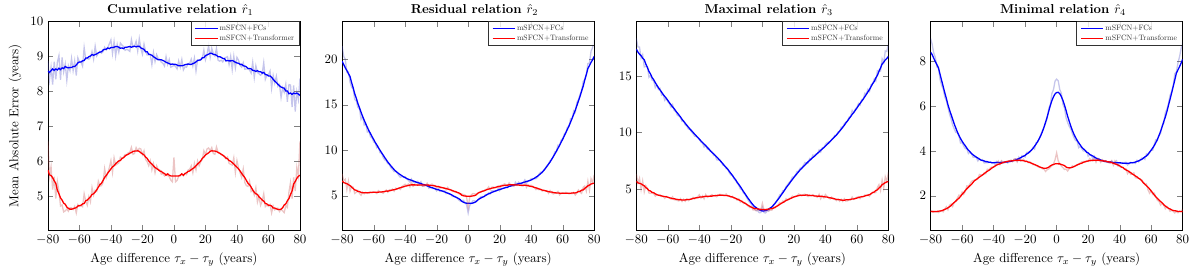}
    \caption{The MAEs of the relation estimation of models with the age difference $\tau_x-\tau_y$ of the pair input image $(x,y)$ when both $x$ and $y$ are sampled from the testing set. (The blue line shows the accuracies of mSFCN$_s$+FCs while the red line shows the accuracies of mSFCN$_s$+Transformer)}
    \label{fig:agedifference}
\end{figure}

Fig.~\ref{fig:agedifference} shows the
 MAEs of relation estimation with the age difference $\tau_x-\tau_y$ of the pair input $(x,y)$ sampled from the testing set.
We only compare the two models: mSFCN$_i$+FC and mSFCN$_i$+Transformer.
It can be seen from the figure that mSFCN$_i$+Transformer provides 
 lower MAE
than mSFCN$_s$+FC and it is less sensitive to the age difference $\tau_x-\tau_y$ than mSFCN$_s$+FC.

\subsubsection{Effects of 1,2,or 4 CNN models for estimating 4 relations}

Table~\ref{tab:differenscenrs} shows accuracies of deep relation learning with the joint, pair and single learning with 1, 2, and 4 CNNs, respectively.
There is no significant difference in results among these three different configurations.
However, joint relation learning only needs 1 neural network to learn the four relations and it requires fewer parameters, memories, and computational times than single relation learning which needs 4 different neural networks, and pair relation learning which needs 2 different neural networks.
In the following sections, we use the estimated relations of joint relation learning for brain age estimation, which only requires to train 1 CNN model for estimating 4 relations.

\begin{table}[!t]
    \centering
    \caption{Accuracy of relation estimation with different configurations.}
    \label{tab:differenscenrs}
    \begin{tabular}{ll|ccc}
    \toprule
    \multicolumn{2}{c|}{Relations}& MAE & CS($\alpha=5$) & Pearson \\
    \midrule
    \multirow{3}{*}{$\hat{r}_1$} &
    Single & 3.95$\pm$0.20 & 73.48\%$\pm$1.91\% & 0.9861$\pm$0.0024\\
    &Pair &  3.85$\pm$0.26 & 73.88\%$\pm$3.00\% & 0.9869$\pm$0.0020 \\
    &Joint & \textbf{3.71$\pm$0.18} &\textbf{75.04\%$\pm$2.01\%} &\textbf{0.9878$\pm$0.0016} \\
    \midrule
    \multirow{3}{*}{$\hat{r}_2$} &
    Single &  3.98$\pm$0.20 & 71.80\%$\pm$2.06\% & 0.9869$\pm$0.0018\\
    &Pair &  3.89$\pm$0.21 & 73.95\%$\pm$2.41\% & 0.9872$\pm$0.0019 \\
    &Joint & \textbf{3.80$\pm$0.24} &\textbf{74.17\%$\pm$2.06\%} &\textbf{0.9880$\pm$0.0018}  \\
    \midrule
    \multirow{3}{*}{$\hat{r}_3$} &
    Single & 3.15$\pm$0.14 & 80.78\%$\pm$0.89\% & 0.9828$\pm$0.0027\\
    &Pair &  \textbf{3.00$\pm$0.14} & \textbf{83.12\%$\pm$1.24\%} & 0.9834$\pm$0.0029  \\
    &Joint & 3.03$\pm$0.16 &81.62\%$\pm$1.44\% &\textbf{0.9838$\pm$0.0027} \\
    \midrule
    \multirow{3}{*}{$\hat{r}_4$} &
    Single & \textbf{1.64$\pm$0.09} & \textbf{94.46\%$\pm$0.60\%} & 0.9804$\pm$0.0067\\
    &Pair &  1.67$\pm$0.12 & 93.99\%$\pm$0.79\% & 0.9805$\pm$0.0079 \\
    &Joint & 1.69$\pm$0.11 &94.19\%$\pm$0.36\% &\textbf{0.9825$\pm$0.0044}\\
    \bottomrule
    \end{tabular}
\end{table}

\begin{figure*}[!t]
    \centering
    \includegraphics[width=\textwidth]{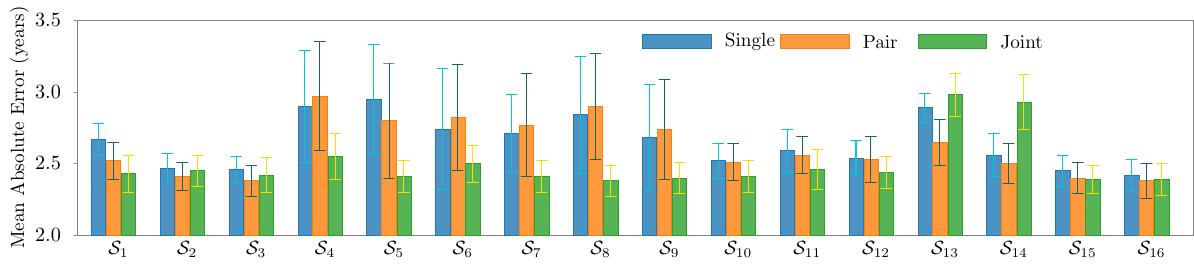}
    \caption{The Mean Absolute Error (MAE) of the joint, pair, and single relation learning with $\mathcal{S}_i$ indexed based on different brain age estimation strategies as shown on Table~\ref{tab:relationsummary}. }
    \label{fig:superbar}
\end{figure*}

\subsection{Accuracy of relation-based brain age estimation}
\label{sec:ac02}

Results in this section correspond to the experiment setting described in Section~\ref{sec:acp02}.
Accuracies of brain age estimation are measured by the three metrics of MAE, CS($\alpha$=5) and Pearson correlation as described in Section~\ref{sec:D1} on the test samples.
Higher accuracy means lower MAE and higher scores of CS and Pearson correlation for brain age estimation in this section.

\subsubsection{Accuracy when $x\neq y$, $x$ and $y$ both from the testing set}

Sub Table I in Table~\ref{tab:relationsummary} shows the accuracies  of the 8 underlying CNN variations for brain age estimation based on pairs of input images sampled from the test set.
In general, models with the shared feature extraction backbones provide better accuracies than the independent backbones and using Transformer for relation regression gives better results than using CNN.
The mSFCN$_i$+Transformer provides the lowest  MAEs and highest scores of CS($\alpha$=5) and Pearson correlation among all models.
The accuracy goes even higher by the ensemble of different relations,  with an MAE of 2.42 years.

\subsubsection{Accuracy on $x$ with reference $y$ when $x\neq y$, $x$ from the testing set and $y$ from the training set}

\begin{figure}[!ht]
    \centering
    \includegraphics[width=0.45\textwidth]{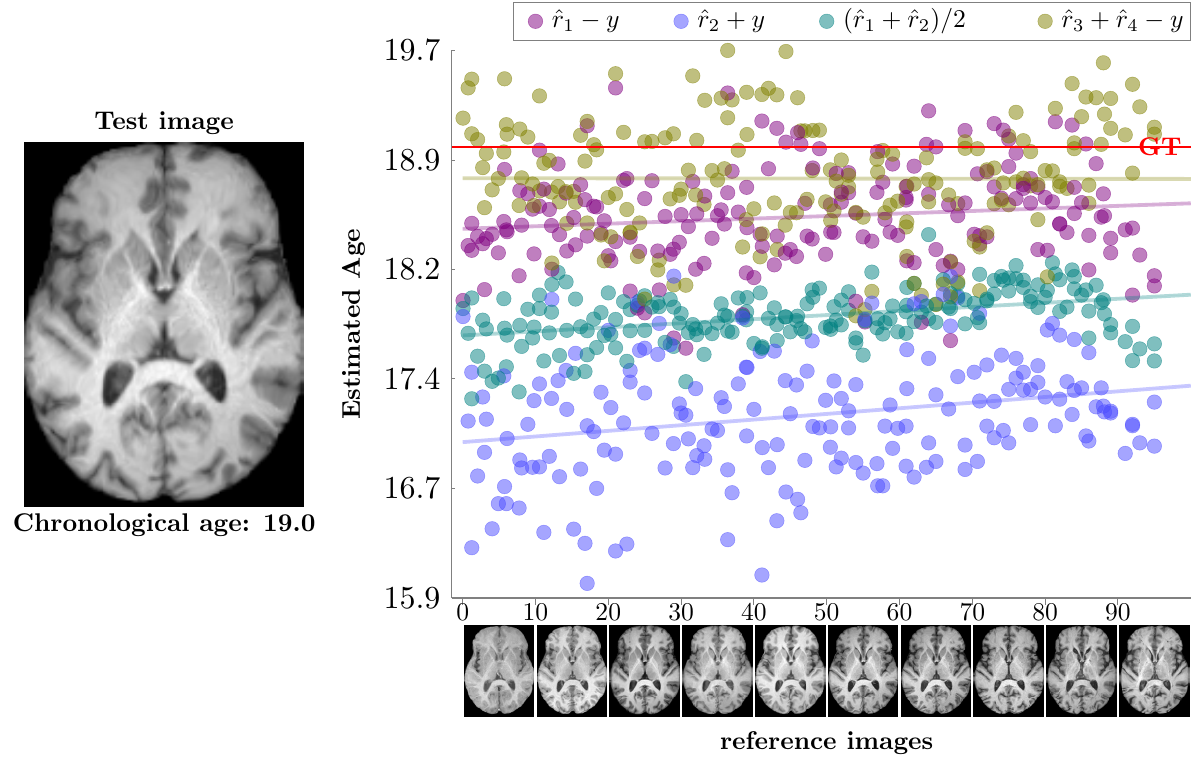}
    \caption{An example of brain age estimation of the mSFCN$_s$+Transformer given the test image $x$ and the reference images $y$ from 0-97 years of age sampled from the training set. Each dot represents the estimated age with one reference image according to the estimated relations $\hat{r}_i,i\in\{1,2,3,4\}$. The red line indicates the chronological age.}
    \label{fig:relexample}
\end{figure}

Fig.~\ref{fig:relexample} shows an example of the age estimation $\hat{\tau}_x$ (x from the testing set whose age is to be estimated) based on different reference images $y$ (from the training set) according to four estimated relations $\hat{r}_i, i\in\{1,2,3,4\}$.
We compute the average age of brain age estimations based on different references $y$ as the final estimated brain age based on learned relations.
The estimated brain ages $\hat{\tau}_x$ are slightly different given different reference images $y$.

\begin{figure*}[!ht]
    \centering
    \includegraphics[width=\textwidth]{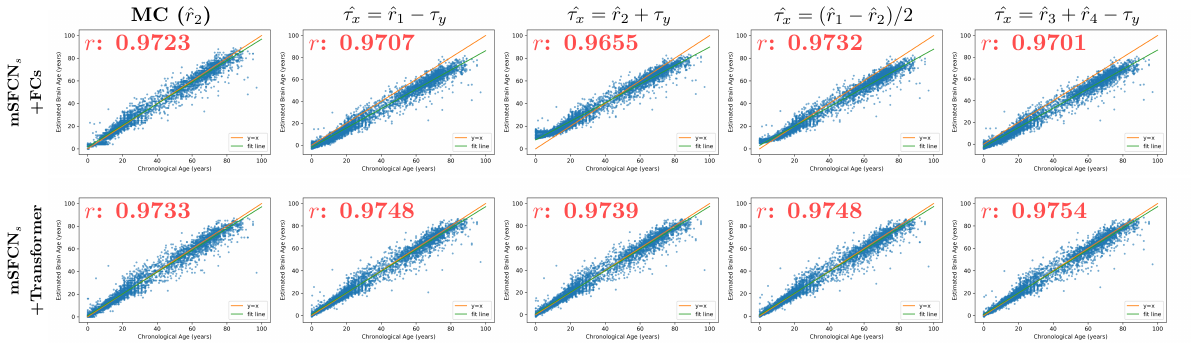}
    \caption{Scatter plots of the estimated and chronological ages based on different relations with the reference $y$ sampled from the training set. The orange lines indicate the ideal estimation when the estimated age equals the chronological age while the green lines are the fitted regression lines. The $r$ is the Spearsman correlation between the estimated brain age and chronological brain age.}
    \label{fig:scatter}
\end{figure*}

Sub Table II in Table~\ref{tab:relationsummary} shows the accuracies
of the brain age estimation $\hat{\tau}_x$ according to the learned relations $\hat{r}_i,i\in\{1,2,3,4\}$ of the pair input $(x,y)$ with the known reference $y$ sampled from the training set.
From the table we can obtain the following observations:
(1) The method using the MC rule~\cite{lim2019order} ( only using the binarized $\hat{r}_2$ for brain age estimation) provides better results than proposed methods when using FCs for the relation regression.
However,  our proposed relation regression based on Transformer gives better results than the method of MC rule. 
In addition, the computation of the MC method~\cite{lim2019order} takes a long time due to its high complexity.
(2) In about $75\%$ of the cases, the shared backbone ($\mathcal{F}_1=\mathcal{F}_2$) for feature extraction gives higher accuracies than the independent backbone ($\mathcal{F}_1\neq\mathcal{F}_2$).
(3) When using the CNN for the relation regression, models using SFCN as the backbone provides lower MAEs than models using mSFCN as the backbone. 
However, when using the Transformer for the relation regression, models using the mSFCN as the backbone have lower MAEs than models using the SFCN as the backbone.
(4) The mSFCN$_s$+Transformer provides lower MAEs than other model variations and the highest accuracy is given by using the ``maximal relation ($\hat{r}_3$)" and ``minimal relation ($\hat{r}_4$)" in terms of MAE and Pearson coefficient, with an MAE of 2.38 years.
Fig.~\ref{fig:scatter} shows the scatter plots between the estimated brain age and chronological age based on different relations.

\begin{table*}[!t]
    \centering
    \caption{The summary of the accuracies for brain age estimation.}
    \label{tab:relationsummary}
    \resizebox{\textwidth}{!}{
    \begin{tabular}{lll|cc|cc|cc|cc}
    \toprule
    \multicolumn{3}{c|}{\multirow{4}{*}{Different strategies: $\mathcal{S}_i$}} & \multicolumn{4}{c|}{Relation regression based on FCs~\cite{lim2019order}} & \multicolumn{4}{c}{Relation regression based on Transformer}\\
    \cmidrule{4-11}
    & & & \multicolumn{2}{c|}{CNN Backbone: SFCN~\cite{peng2019accurate}} & \multicolumn{2}{c|}{CNN Backbone: mSFCN} & \multicolumn{2}{c|}{CNN Backbone: SFCN~\cite{peng2019accurate}} & \multicolumn{2}{c}{CNN Backbone: mSFCN}  \\
    \cmidrule{4-11}
    & & & Shared  & Independent  & Shared & Independent  & Shared  & Independent & Shared  & Independent  \\
    & & & $\mathcal{F}_1=\mathcal{F}_2$ &  $\mathcal{F}_1\neq\mathcal{F}_2$ &  $\mathcal{F}_1=\mathcal{F}_2$ &  $\mathcal{F}_1\neq\mathcal{F}_2$ &  $\mathcal{F}_1=\mathcal{F}_2$ &  $\mathcal{F}_1\neq\mathcal{F}_2$&  $\mathcal{F}_1=\mathcal{F}_2$ &  $\mathcal{F}_1\neq\mathcal{F}_2$ \\
\midrule
 \multicolumn{11}{l}{\CC Sub Table I: The accuracies of brain age estimation with the pair inputs $x\neq y$ sampled from the test set. The accuracies of the $\hat{\tau_x}$ and $\hat{\tau_y}$ are measured together for brain age estimation. } \\
\midrule
\multirow{3}{*}{$\mathcal{S}_1$} & \multirow{3}{*}{\parbox{2.5cm}{$\hat{\tau}_x=(\hat{r}_1+\hat{r}_2)/2$ \\ $\hat{\tau}_y=(\hat{r}_1-\hat{r}_2)/2$ }} & MAE & 3.57$\pm$0.21 & 3.62$\pm$0.15 & 3.85$\pm$0.33 & 3.71$\pm$0.26 & 3.31$\pm$0.09 & 3.19$\pm$0.23 & 2.43$\pm$0.13 & 2.54$\pm$0.07 \\
&& CS($\alpha=5$)& 76.53\%$\pm$2.57\% & 75.84\%$\pm$2.25\% & 72.83\%$\pm$4.07\% & 74.67\%$\pm$2.82\% & 78.91\%$\pm$1.33\% & 80.40\%$\pm$3.09\% & 87.37\%$\pm$0.62\% & 86.33\%$\pm$0.61\% \\
&& Pearson& 0.9837$\pm$0.0007 & 0.9825$\pm$0.0015 & 0.9860$\pm$0.0016 & 0.9836$\pm$0.0030 & 0.9800$\pm$0.0014 & 0.9811$\pm$0.0023 & 0.9879$\pm$0.0017 & 0.9867$\pm$0.0015\\
\midrule
\multirow{3}{*}{$\mathcal{S}_2$} &
\multirow{3}{*}{\parbox{3.5cm}{$\hat{\tau}_x=\hat{r}_3,\hat{\tau}_y=\hat{r}_4 \ \ \ \text{if} \ \ \ \hat{r}_2>0$ \\ $\hat{\tau}_x=\hat{r}_4,\hat{\tau}_y=\hat{r}_3 \ \ \ $ Otherwise }} & MAE & 4.46$\pm$0.24 & 4.97$\pm$0.34 & 4.56$\pm$0.40 & 4.57$\pm$0.31 & 2.98$\pm$0.10 & 3.17$\pm$0.11 & 2.45$\pm$0.11 & 2.58$\pm$0.08\\
&& CS($\alpha=5$)&  65.63\%$\pm$1.76\% & 59.99\%$\pm$3.67\% & 66.83\%$\pm$3.12\% & 66.00\%$\pm$3.03\% & 82.23\%$\pm$0.83\% & 80.85\%$\pm$1.76\% & 87.19\%$\pm$0.92\% & 85.98\%$\pm$0.86\%\\
&& Pearson& 0.9770$\pm$0.0015 & 0.9705$\pm$0.0036 & 0.9813$\pm$0.0011 & 0.9768$\pm$0.0031 & 0.9815$\pm$0.0019 & 0.9808$\pm$0.0010 & 0.9878$\pm$0.0017 & 0.9857$\pm$0.0019\\
\midrule
\multirow{3}{*}{$\mathcal{S}_3$} &
\multirow{3}{*}{\parbox{3.5cm}{Ensemble}} & MAE & 3.92$\pm$0.21 & 4.14$\pm$0.22 & 4.13$\pm$0.36 & 4.05$\pm$0.29 & 3.07$\pm$0.10 & 3.13$\pm$0.17 & 2.42$\pm$0.12 & 2.53$\pm$0.08\\
&& CS($\alpha=5$)&  72.18\%$\pm$1.89\% & 69.27\%$\pm$2.80\% & 70.39\%$\pm$3.64\% & 71.47\%$\pm$2.96\% & 81.38\%$\pm$0.92\% & 81.06\%$\pm$2.28\% & 87.34\%$\pm$0.90\% & 86.41\%$\pm$0.77\%\\
&& Pearson& 0.9817$\pm$0.0009 & 0.9790$\pm$0.0019 & 0.9847$\pm$0.0012 & 0.9816$\pm$0.0032 & 0.9815$\pm$0.0015 & 0.9815$\pm$0.0015 & 0.9879$\pm$0.0016 & 0.9863$\pm$0.0017\\
\midrule
\multicolumn{11}{l}{\CC Sub Table II: The accuracies of different reference methods to estimate the $\hat{\tau_x}$ according to the learned relations $\hat{r}_i$ with the references $y$ sampled from the training samples with known age information. } \\
\midrule
\multirow{3}{*}{$\mathcal{S}_4$} &
\multirow{3}{*}{MC~\cite{lim2019order} ($\hat{r}_2$)} & MAE & 2.99$\pm$0.14 & 3.75$\pm$0.30 & 2.73$\pm$0.04 & 3.06$\pm$0.22 & 5.27$\pm$0.24 & 3.82$\pm$0.47 & 2.55$\pm$0.16 & 2.61$\pm$0.15 \\
&& CS($\alpha=5$)& 84.48\%$\pm$1.87\% & 75.06\%$\pm$3.14\% & 87.58\%$\pm$0.54\% & 83.90\%$\pm$3.17\% & 64.99\%$\pm$1.66\% & 75.84\%$\pm$4.46\% & 87.67\%$\pm$1.15\% & 87.31\%$\pm$1.45\%\\
&& Pearson& 0.9859$\pm$0.0013 & 0.9827$\pm$0.0016 & 0.9876$\pm$0.0014 & 0.9864$\pm$0.0017 & 0.9475$\pm$0.0048 & 0.9731$\pm$0.0062 & 0.9871$\pm$0.0018 & 0.9860$\pm$0.0024\\
\midrule
\multirow{3}{*}{$\mathcal{S}_5$} &
\multirow{3}{*}{$\hat{\tau}_x=\hat{r}_1-\tau_y$} & MAE & 4.06$\pm$0.60 & 3.77$\pm$0.55 & 5.96$\pm$1.18 & 4.80$\pm$0.95 & 2.63$\pm$0.10 & 3.54$\pm$0.36 & 2.41$\pm$0.11 & 2.59$\pm$0.12\\
&& CS($\alpha=5$)& 71.06\%$\pm$7.54\% & 73.83\%$\pm$8.38\% & 47.94\%$\pm$12.96\% & 62.77\%$\pm$9.09\% & 85.57\%$\pm$0.76\% & 76.97\%$\pm$3.39\% & 87.26\%$\pm$0.84\% & 86.30\%$\pm$1.09\%\\
&& Pearson& 0.9859$\pm$0.0010 & 0.9853$\pm$0.0006 & 0.9877$\pm$0.0017 & 0.9867$\pm$0.0020 & 0.9866$\pm$0.0014 & 0.9775$\pm$0.0045 & 0.9879$\pm$0.0017 & 0.9863$\pm$0.0019\\
\midrule
\multirow{3}{*}{$\mathcal{S}_6$} &
\multirow{3}{*}{$\hat{\tau}_x=\hat{r}_2+\tau_y$} & MAE & 3.90$\pm$0.44 & 4.31$\pm$0.68 & 4.48$\pm$0.71 & 4.39$\pm$0.51 & 4.23$\pm$0.14 & 3.72$\pm$0.44 & 2.50$\pm$0.13 & 2.56$\pm$0.14\\
&& CS($\alpha=5$)& 71.43\%$\pm$4.22\% & 66.68\%$\pm$6.46\% & 65.30\%$\pm$7.50\% & 65.63\%$\pm$6.00\% & 71.26\%$\pm$1.59\% & 75.62\%$\pm$4.35\% & 86.98\%$\pm$1.05\% & 86.43\%$\pm$1.57\%\\
&& Pearson&0.9834$\pm$0.0014 & 0.9824$\pm$0.0014 & 0.9855$\pm$0.0024 & 0.9858$\pm$0.0023 & 0.9653$\pm$0.0035 & 0.9737$\pm$0.0061 & 0.9874$\pm$0.0018 & 0.9864$\pm$0.0025\\
\midrule
\multirow{3}{*}{$\mathcal{S}_7$} &
\multirow{3}{*}{$\hat{\tau}_x=(\hat{r}_1-\hat{r}_2)/2$} & MAE & 3.57$\pm$0.28 & 3.67$\pm$0.16 & 3.74$\pm$0.25 & 3.82$\pm$0.27 & 3.30$\pm$0.07 & 3.58$\pm$0.39 & 2.41$\pm$0.11 & 2.55$\pm$0.13\\
&& CS($\alpha=5$)& 73.24\%$\pm$3.56\% & 72.24\%$\pm$1.94\% & 73.67\%$\pm$5.15\% & 69.70\%$\pm$3.44\% & 79.01\%$\pm$0.86\% & 76.43\%$\pm$3.86\% & 87.33\%$\pm$0.71\% & 86.44\%$\pm$1.28\%\\
&& Pearson& 0.9849$\pm$0.0008 & 0.9840$\pm$0.0011 & 0.9872$\pm$0.0017 & 0.9864$\pm$0.0021 & 0.9788$\pm$0.0018 & 0.9760$\pm$0.0051 & 0.9878$\pm$0.0017 & 0.9864$\pm$0.0022\\
\midrule
\multirow{3}{*}{$\mathcal{S}_8$} &
\multirow{3}{*}{$\hat{\tau}_x=\hat{r}_3+\hat{r}_4-\tau_y$} & MAE & 4.17$\pm$0.56 & 3.75$\pm$0.61 & 6.17$\pm$1.35 & 4.92$\pm$0.96 & 2.60$\pm$0.10 & 3.42$\pm$0.22 & 2.38$\pm$0.11 & 2.56$\pm$0.14\\
&& CS($\alpha=5$)& 67.94\%$\pm$7.14\% & 74.16\%$\pm$10.03\% & 43.53\%$\pm$15.36\% & 61.71\%$\pm$9.32\% & 85.59\%$\pm$0.84\% & 78.35\%$\pm$2.06\% & 87.40\%$\pm$0.80\% & 85.87\%$\pm$1.39\%\\
&& Pearson& 0.9861$\pm$0.0011 & 0.9855$\pm$0.0006 & 0.9875$\pm$0.0016 & 0.9866$\pm$0.0020 & 0.9861$\pm$0.0018 & 0.9754$\pm$0.0073 & 0.9881$\pm$0.0017 & 0.9860$\pm$0.0024\\
\midrule
\multirow{3}{*}{$\mathcal{S}_9$} &
\multirow{3}{*}{Ensemble} & MAE & 3.66$\pm$0.36 & 3.60$\pm$0.12 & 4.04$\pm$0.38 & 3.96$\pm$0.39 & 3.13$\pm$0.08 & 3.50$\pm$0.34 & 2.40$\pm$0.11 & 2.53$\pm$0.13\\
&& CS($\alpha=5$)& 73.50\%$\pm$4.52\% & 73.42\%$\pm$2.19\% & 72.29\%$\pm$3.87\% & 69.90\%$\pm$5.06\% & 80.42\%$\pm$0.81\% & 77.32\%$\pm$3.22\% & 87.31\%$\pm$0.84\% & 86.42\%$\pm$1.36\%\\
&& Pearson& 0.9853$\pm$0.0007 & 0.9849$\pm$0.0008 & 0.9874$\pm$0.0019 & 0.9867$\pm$0.0020 & 0.9808$\pm$0.0018 & 0.9763$\pm$0.0051 & 0.9879$\pm$0.0017 & 0.9864$\pm$0.0022\\
\midrule
\multicolumn{11}{l}{\CC Sub Table III: The accuracies of brain age estimation $\hat{\tau_x}$ when the pair inputs are the same $y=x$. } \\
\midrule
\multirow{3}{*}{$\mathcal{S}_{10}$} &
\multirow{3}{*}{$\hat{\tau}_x=\hat{r}_1/2$} & MAE & 3.11$\pm$0.17 & 3.05$\pm$0.11& 3.27$\pm$0.24& 3.18$\pm$0.24& 2.71$\pm$0.12 & 2.84$\pm$0.18& 2.41$\pm$0.11& 2.40$\pm$0.08\\
&& CS($\alpha=5$)& 81.37\%$\pm$2.24\% & 82.04\%$\pm$1.60\%& 78.79\%$\pm$3.41\%& 79.27\%$\pm$2.92\%& 84.81\%$\pm$0.86\% & 83.52\%$\pm$1.85\%& 87.28\%$\pm$0.89\%& 87.35\%$\pm$0.94\%\\
&& Pearson& 0.9867$\pm$0.0010 & 0.9868$\pm$0.0013& 0.9881$\pm$0.0019& 0.9880$\pm$0.0021& 0.9861$\pm$0.0014 & 0.9852$\pm$0.0014& 0.9879$\pm$0.0017& 0.9880$\pm$0.0015\\
\midrule
\multirow{3}{*}{$\mathcal{S}_{11}$} &
\multirow{3}{*}{$\hat{\tau}_x=(\hat{r}_1+\hat{r}_2)/2$} & MAE & 3.09$\pm$0.17 & 3.07$\pm$0.19& 3.33$\pm$0.24& 3.26$\pm$0.36& 2.69$\pm$0.05 & 2.85$\pm$0.11& 2.46$\pm$0.14& 2.52$\pm$0.03\\
&& CS($\alpha=5$)& 81.57\%$\pm$2.22\% & 82.02\%$\pm$2.49\%& 78.60\%$\pm$3.60\%& 79.13\%$\pm$4.22\%& 85.67\%$\pm$0.93\% & 84.07\%$\pm$1.35\%& 87.46\%$\pm$0.91\%& 86.75\%$\pm$0.55\%\\
&& Pearson& 0.9865$\pm$0.0009 & 0.9862$\pm$0.0016& 0.9879$\pm$0.0019& 0.9870$\pm$0.0027& 0.9869$\pm$0.0014 & 0.9852$\pm$0.0004& 0.9877$\pm$0.0017& 0.9870$\pm$0.0013\\
\midrule
\multirow{3}{*}{$\mathcal{S}_{12}$} &
\multirow{3}{*}{$\hat{\tau}_x=(\hat{r}_1-\hat{r}_2)/2$} & MAE & 3.16$\pm$0.19 & 3.26$\pm$0.20& 3.24$\pm$0.25& 3.24$\pm$0.28& 3.55$\pm$0.13 & 3.65$\pm$0.45& 2.44$\pm$0.11& 2.56$\pm$0.15\\
&& CS($\alpha=5$)& 80.45\%$\pm$2.38\% & 79.32\%$\pm$1.85\%& 78.82\%$\pm$3.34\%& 78.37\%$\pm$3.31\%& 77.21\%$\pm$1.03\% & 76.10\%$\pm$3.93\%& 87.28\%$\pm$0.81\%& 86.44\%$\pm$1.48\%\\
&& Pearson& 0.9868$\pm$0.0011 & 0.9861$\pm$0.0012& 0.9882$\pm$0.0020& 0.9878$\pm$0.0018& 0.9767$\pm$0.0024 & 0.9757$\pm$0.0053& 0.9878$\pm$0.0017& 0.9864$\pm$0.0022\\
\midrule
\multirow{3}{*}{$\mathcal{S}_{13}$} &
\multirow{3}{*}{$\hat{\tau}_x=\hat{r}_3$} & MAE & 3.91$\pm$0.84 & 4.91$\pm$0.59& 3.04$\pm$0.48&
3.84$\pm$0.32& 3.60$\pm$0.23 & 3.86$\pm$0.15& 2.98$\pm$0.15& 3.20$\pm$0.19\\
&& CS($\alpha=5$)& 72.44\%$\pm$9.74\% & 58.97\%$\pm$6.93\%& 82.13\%$\pm$5.72\%& 73.49\%$\pm$4.53\%& 76.69\%$\pm$2.79\% & 74.01\%$\pm$2.78\%& 84.59\%$\pm$1.27\%& 81.95\%$\pm$1.72\%\\
&& Pearson& 0.9865$\pm$0.0012 & 0.9865$\pm$0.0013& 0.9881$\pm$0.0018& 0.9872$\pm$0.0020& 0.9832$\pm$0.0017 & 0.9833$\pm$0.0012& 0.9879$\pm$0.0015& 0.9873$\pm$0.0018\\
\midrule
\multirow{3}{*}{$\mathcal{S}_{14}$} &
\multirow{3}{*}{$\hat{\tau}_x=\hat{r}_4$} & MAE & 4.98$\pm$0.96 & 6.23$\pm$0.27& 5.32$\pm$0.65& 6.17$\pm$0.69& 3.27$\pm$0.20 & 3.25$\pm$0.16& 2.93$\pm$0.19& 2.76$\pm$0.13\\
&& CS($\alpha=5$)& 61.25\%$\pm$10.55\% & 47.37\%$\pm$3.64\%& 58.24\%$\pm$6.22\%& 49.23\%$\pm$8.67\%& 80.72\%$\pm$1.84\% & 81.45\%$\pm$1.56\%& 84.70\%$\pm$1.03\%& 85.51\%$\pm$1.37\%\\
&& Pearson& 0.9863$\pm$0.0010 & 0.9858$\pm$0.0011& 0.9872$\pm$0.0020& 0.9870$\pm$0.0024& 0.9846$\pm$0.0017 & 0.9814$\pm$0.0056& 0.9877$\pm$0.0018& 0.9871$\pm$0.0024\\
\midrule
\multirow{3}{*}{$\mathcal{S}_{15}$} &
\multirow{3}{*}{$\hat{\tau}_x=(\hat{r}_3+\hat{r}_4)/2$} & MAE & 3.05$\pm$0.13 & 2.95$\pm$0.13& 3.13$\pm$0.26& 3.09$\pm$0.22& 2.75$\pm$0.09 & 2.90$\pm$0.13& 2.39$\pm$0.10 & 2.44$\pm$0.09\\
&& CS($\alpha=5$)& 81.70\%$\pm$1.81\% & 83.40\%$\pm$1.79\%& 80.99\%$\pm$3.95\% & 80.38\%$\pm$2.76\%& 84.44\%$\pm$0.90\% & 83.45\%$\pm$1.53\%& 87.51\%$\pm$0.85\% & 87.32\%$\pm$0.99\%\\
&& Pearson& 0.9867$\pm$0.0010 & 0.9868$\pm$0.0013& 0.9881$\pm$0.0019& 0.9880$\pm$0.0021& 0.9852$\pm$0.0015 & 0.9841$\pm$0.0024& 0.9880$\pm$0.0016& 0.9877$\pm$0.0019\\
\midrule
\multirow{3}{*}{$\mathcal{S}_{16}$} &
\multirow{3}{*}{Ensemble} & MAE & 3.07$\pm$0.14 & 2.99$\pm$0.12& 3.20$\pm$0.25& 3.14$\pm$0.22& 2.71$\pm$0.10 & 2.85$\pm$0.15& 2.39$\pm$0.11& 2.41$\pm$0.08\\
&& CS($\alpha=5$)& 81.63\%$\pm$1.98\% & 82.47\%$\pm$1.52\%& 79.97\%$\pm$3.74\%& 79.84\%$\pm$2.84\%& 84.57\%$\pm$0.84\% & 83.62\%$\pm$1.63\%& 87.33\%$\pm$0.90\%& 87.29\%$\pm$0.89\%\\
&& Pearson& 0.9867$\pm$0.0010 & 0.9868$\pm$0.0013& 0.9881$\pm$0.0019& 0.9880$\pm$0.0021& 0.9857$\pm$0.0014 & 0.9849$\pm$0.0016& 0.9880$\pm$0.0016& 0.9879$\pm$0.0017\\
\bottomrule
\end{tabular}}
\end{table*}

\subsubsection{Accuracy when $x=y$ and both from the testing set}

\begin{figure}[!t]
    \centering
    \includegraphics[width=0.5\textwidth]{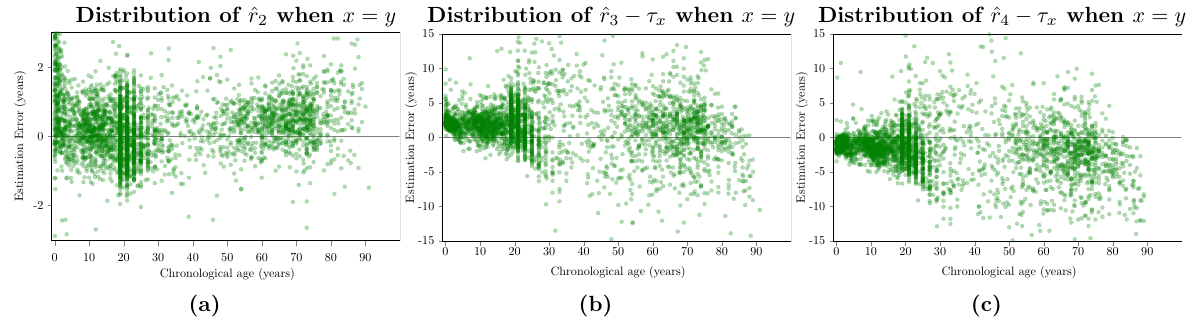}
    \caption{The scatter plots of the $\hat{r}_2$ (a), $\hat{r}_3-\tau_x$ (b) and $\hat{r}_4-\tau_x$ (c) versus chronological age when the pair input is the same: $y=x$. These three figures show the regression errors from the mSFCN$_s$+Transformer model for relation learning.}
    \label{fig:errdistribution}
\end{figure}

In this section, we feed the neural network with the same image as the pair of input: $y=x$.
Ideally, when $y=x$, the ``relative relation $r_2$" should be zero, the ``maximal relation $r_3$" and the ``minimal relation $r_4$" should be equal to the age of input $x$.
In practice, the estimated relations $\hat{r}_2\neq0$ and $\hat{r}_3\neq\hat{r}_4\neq \tau_x$ due to the regression errors from the neural network.
To learn the relations between the pair input samples, the training pair images are randomly sampled ($x\neq y$).
The distributions of these three estimated relations over different chronological ages are shown in Fig.~\ref{fig:errdistribution}.
The estimated error of the $\hat{r}_2$ represents the uncertainty of the model for brain age estimation.
From Fig.~\ref{fig:errdistribution} (b) we can see that the estimation of the ``maximal relation $\hat{r}_3$" is greater than the input $\tau_x$ in most testing samples.
Similarly, from Fig.~\ref{fig:errdistribution} (c) we can see that the estimation of the ``minimal relation $\hat{r}_4$" is smaller than the input $\tau_x$ in most testing samples.

Sub Table III in Table~\ref{tab:relationsummary} shows the accuracies of different model variations for brain age estimation $\hat{\tau}_x$ when $y=x$. 
The lowest  MAEs are from the $\hat{\tau}_x=(\hat{r}_3+\hat{r}_4)/2$ on different CNN model variations. 
In addition, using Transformer provides better results than using CNN for relation regression.
The highest accuracies are given by the mSFCN$_s$+Transformer in terms of MAE and CS score.
We also find that combining these six estimations $\sum_i^6(\hat{\tau}_x)/6$ (ensemble by averaging) does not improve the accuracies.

\subsubsection{Robustness/uncertainty in relation-based brain age estimation}

Fig.~\ref{fig:superbar} shows the MAEs of the joint, pair, and single relation learning  for brain age estimation with different strategies $\mathcal{S}_i$ indexed in Table~\ref{tab:relationsummary}.
In most cases, the joint relation learning provides the highest accuracy, except for the $\mathcal{S}_2$, $\mathcal{S}_3$, $\mathcal{S}_{13}$ and $\mathcal{S}_{14}$ where the relations $r_3$ and $r_4$ are involved in the computation.
Table~\ref{tab:relationsummary} and Fig.~\ref{fig:superbar} show that the best result of brain age estimation is from the strategy of $\mathcal{S}_8$ and $\mathcal{S}_{15}$.

\begin{figure*}[!t]
    \centering
    \includegraphics[width=\textwidth]{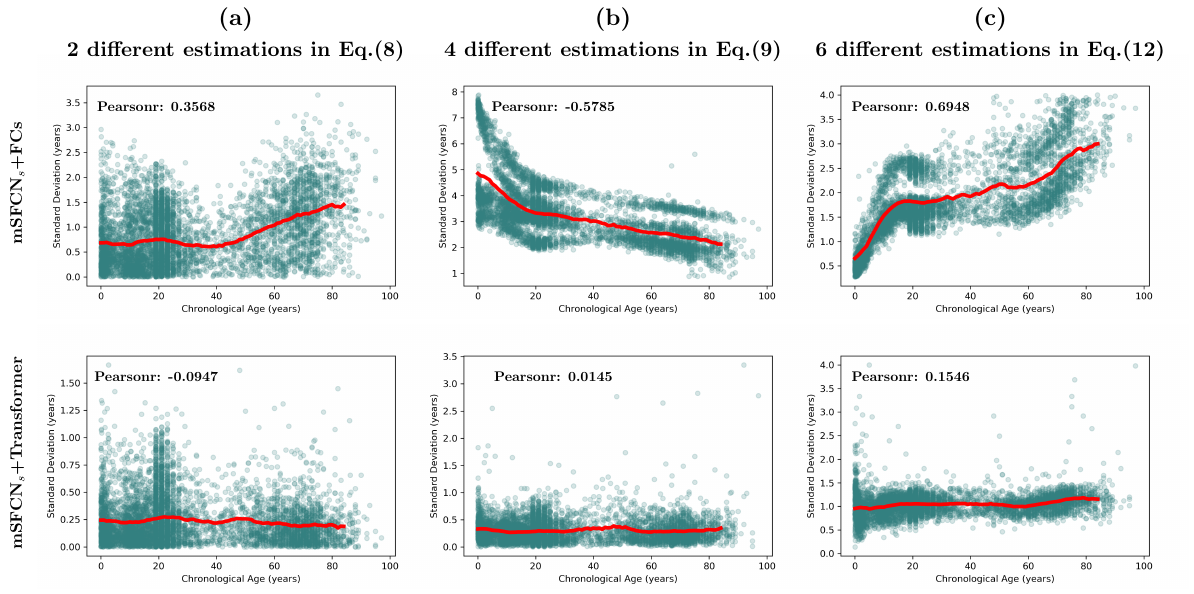}
    \caption{The uncertainty which is the standard deviation of estimated brain age from the different estimations in Eq.~\ref{eq:xny} (Figure (a)), Eq.~\ref{eq:e10} (Figure (b)) and Eq.~\ref{eq:selfcom} (Figure (c)) over different chronological ages. The Pearson correlation is computed between the uncertainty and chronological age on all test subjects. The red lines are the average uncertainty over years.}
    \label{fig:uncertainty}
\end{figure*}

Different brain age estimations can be obtained according to Eqs.~\ref{eq:xny},~\ref{eq:e10} and~\ref{eq:selfcom} and the uncertainty can be measured as the standard deviation of these different brain age estimations~\cite{hepp2021uncertainty,palma2020quantifying}, which shows how uncertainty for brain age estimation can be introduced by the deep relation learning.
Fig.~\ref{fig:uncertainty} shows the distribution of the uncertainty over different chronological ages. 
Uncertainty computed from different estimations in Eqs.~\ref{eq:xny},~\ref{eq:e10} and~\ref{eq:selfcom} corresponds to three different brain age estimation methods.
The mSFCN$_s$+Transformer has almost the same uncertainty on different ages and the Pearson correlations between the uncertainty and chronological age are smaller than the mSFCN$_s$+FCs on these three different estimation methods.

\subsubsection{Accuracy comparison with state-of-the-art brain age estimation algorithms}

\begin{table*}[!t]
    \centering
    \caption{The comparison of accuracies of different methods on the discovery cohort with 5-fold cross-validation.}
    \label{tab:soacomparison}
    \begin{tabular}{l|lcc}
    \toprule
    Method &  MAE & CS($\alpha=5$) & Pearson \\
    \midrule
    $^1$ Hi-Net~\cite{zhou2020hi} & 3.12$\pm$0.22****& 80.5\% & 0.983\\
    $^1$ FiA-Net~\cite{he2021multi}  & 3.00$\pm$0.06**** & 82.1\% & 0.984\\
    3D CNN in~\cite{cole2017predicting}  & 3.93$\pm$0.56**** &  71.26\%$\pm$6.98\% &  0.981$\pm$0.002 \\
    GL-Transformer~\cite{he2021global}  & 2.59$\pm$0.18*** & 85.83\%$\pm$1.45\% & 0.987$\pm$0.001\\
    SFCN~\cite{peng2019accurate}  & 2.62$\pm$0.07**** & 84.47\%$\pm$0.79\% & 0.987$\pm$0.002\\
    2D DeepBrainNet~\cite{brainwaa160}  & 2.59$\pm$0.09*** & 86.45\%$\pm$0.39\% & 0.985$\pm$0.003 \\
    \midrule
    mSFCN   &  2.64$\pm$0.17**** & 85.58\%$\pm$1.28\% &  0.986$\pm$0.003 \\
    mSFCN+Transformer  & 2.56$\pm$0.07** & 85.92\%$\pm$1.13\% &  0.987$\pm$0.001 \\
    $\S$ mSFCN$_s$+Transformer+Relation(with $y=x$) & 2.39$\pm$0.11 & 87.33\%$\pm$0.90\% & \textbf{0.988$\pm$0.002}\\
    $\S$ mSFCN$_s$+Transformer+Relation(with $y\neq x$) & \textbf{2.38$\pm$0.11} & \textbf{87.40\%$\pm$0.80\%} & \textbf{0.988$\pm$0.002}\\
    \bottomrule
    \multicolumn{4}{l}{$1:$ Results are from the work~\cite{he2021multi}. The methods are based on 3D CNN.}\\
    \multicolumn{4}{l}{$\S$ The best results of the proposed deep relation learning in different configurations. }\\
    \multicolumn{4}{l}{Significance level: *($p<$0.05), **:($p<$0.01), ***:($p<$0.001),****:($p<$0.0001) }\\
    \multicolumn{4}{l}{Significance is measured by using paired $t$-test (two-side) based on the absolute errors across all test samples. }\\
    \bottomrule
    \end{tabular}
\end{table*}

Table~\ref{tab:soacomparison} shows the accuracies of different state-of-the-art models for brain age estimation.
With the same training dataset and configuration, the 2D DeepBrainNet~\cite{brainwaa160} gives slightly higher accuracies than 3D SFCN~\cite{peng2019accurate}.
The mSFCN model, which is similar to SFCN~\cite{peng2019accurate} but has the max-pooling layer at the beginning, has a similar accuracy with the SFCN.
However, mSFCN is more efficient than SFCN since it reduces the image size at the beginning using max-pooling.
Our proposed deep relation learning with reference image $y$ or with the same input pair $y=x$ provides higher accuracies than all other models.
The MAE of the proposed method (2.38 years) is lower than MAEs of other models. 
The statistical significance measured by the paired $t$-test (two-side) based on absolute errors across all 6,049 test samples indicates that the differences of the MAEs between the proposed method and other models are statistically significant ($p<0.05$).
In addition, using different reference images $y$ can provide slightly lower MAE and higher CS score than using the same input pair when $y=x$.
The experimental results show that the combination of the mSFCN and Transformer can provide better results than other models.
Moreover, the neural network with deep relation learning (mSFCN+Transformer+Relations) provides lower MAE and higher scores of CS and Pearson correlation than the neural network without relation learning (mSFCN+Transformer), demonstrating that using the four different relations can further reduce the error for brain age estimation.

\begin{figure*}[!t]
    \centering
    \includegraphics[width=\textwidth]{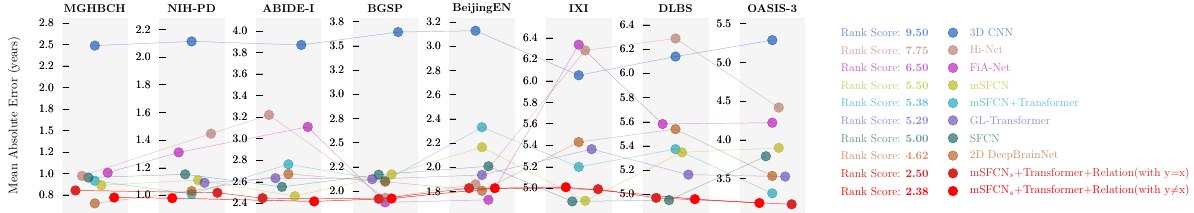}
    \caption{The MAE of different models on each data set. The ranking score is the average rank of each model on the 8 different datasets involved in the cross-validation.}
    \label{fig:subsets}
\end{figure*}

Fig.~\ref{fig:subsets} shows the MAE of different models on the 8 different datasets involved in the cross-validation.
We first rank the models on each dataset and assign a rank score (rank=1,2,...,10 for 10 different models) and the average rank score of each model on the 8 datasets is computed and shown in Fig.~\ref{fig:subsets}.
The result shows that our proposed deep relation learning gives the lowest rank score, demonstrating that deep relation learning provides good generalization on different datasets.

\section{Discussion and Conclusion}
\label{sec:conclusion}

We proposed a deep relation learning for regression given a pair of input images and evaluated it for brain age estimation on brain MRIs.
Four non-linearly correlated relations were evaluated, including the ``cumulative relation", ``relative relation", ``maximal relation" and ``minimal relation".
To learn these relations on a pair of input images, we used a neural network with two parts: feature extraction which was based on convolutional neural networks and relation regression which was based on Transformer.
We evaluated the accuracies of the proposed deep relation learning on a merged dataset with 6,049 healthy brain MRIs acquired between 0-97 years of age.

Our experimental results demonstrate the advantages of relation learning for brain age estimation.
(1) Our proposed relation learning is an extension of the order learning~\cite{lim2019order} and experimental results in Table~\ref{tab:relationsummary} shows that the MAE (2.38 years) of using the four different relations is lower than the MAE (2.55 years) of the order learning based on the MC rules~\cite{lim2019order}.
(2) We evaluated different strategies to estimate the brain ages of a pair of input images: brain age estimation with different test images (strategies $\mathcal{S}_{1-3}$ in Table~\ref{tab:relationsummary}), brain age estimation with the reference with known age (strategies $\mathcal{S}_{4-9}$ in Table~\ref{tab:relationsummary}) and brain age estimation with a pair of the same image  (strategies $\mathcal{S}_{10-16}$ in Table~\ref{tab:relationsummary}).
The accuracies of these different strategies are similar and the lowest MAE is 2.38 years which is better than other state-of-the-art models for brain age estimation (as shown in Table~\ref{tab:soacomparison}).
(3) We also evaluated three different training configurations to learn the four relations: joint, pair, and single learning. Experimental results in Table~\ref{tab:differenscenrs} and Fig.~\ref{fig:superbar} show that the MAEs and the scores of CS($\alpha$=5) and Pearson correlation of these three different learning strategies are similar.
However, joint relation learning only needs 1 neural network to learn the 4 relations simultaneously which requires fewer parameters, memories, and computational times than the pair and single relation learning.
(4) The uncertainty shown in Fig.~\ref{fig:uncertainty} indicates that our proposed method is robust for brain age estimation on different ages.

Fig.~\ref{fig:scatter} also shows the bias problem known as ``regression to the mean": the ages of the older subjects being under-estimated while the ages of the younger subjects being over-estimated.
The regression to the mean (RTM) problem~\cite{barnett2005regression} is a natural statistical phenomenon for many regression problems, and age estimation is not an exception.
This bias, or RTM problem, also exists in other age estimation studies that focus purely on adults, where bias correction has been proposed~\cite{liang2019investigating,de2020commentary,beheshti2019bias}.
Bias correction, however, is also with controversies, for which model should be optimal for correction, how to best quantify bias, and how to best evaluate the effects of correction~\cite{butler2021pitfalls}.
Bias correction is also our ongoing work.

The unbalanced age distribution introduces the bias of relation learning between the pair of input images with different age ranges.
As shown in Fig.~\ref{fig:agedifference}, the errors of relation $r_1$ increase among subjects with age differences between 20 to 40. 
This is also the same for the relations $r_3$ and $r_4$.
In addition, the accuracy of the relation $r_4$ is higher than other relations and one possible reason is that the number of young subjects is larger than the number of old subjects.
A balanced age distribution may mitigate this problem.
Our future work will dive into the data imbalanced problem for relation learning and brain age estimation across the lifespan.

Table~\ref{tab:relationsummary} shows that there is no significant difference among the accuracies of deep relation learning with different strategies (the pair of input images are the same image (x=y) or different images (x$\neq$y)).  
The results demonstrate the consistency and generality of the proposed method.
For consistency, our proposed method provides consistent accuracies with the same image or different images as a pair of input images.
For the generality, our proposed method can be used in different scenarios and it can predict the brain age based on reference images if they are available or based on itself if reference images are not available.

Experimental results in Table~\ref{tab:relationsummary} show that the lowest MAEs and highest CS and Pearson correlations are achieved based on the relations $r_3$ and $r_4$, demonstrating that the neural network can capture non-linear relations better than the linear relations.
In addition, joint relation learning can provide slightly  lower MAEs ($\mathcal{S}_8$ and $\mathcal{S}_{15}$) than single and pair relation learning.

The limitations and future works include: (1) We only focused on the evaluation of the proposed deep relation learning and we did not apply it to potential applications, such as building different chains~\cite{lim2019order} for subjects with different sex, race, ethnicity, etc.
Our proposed method can be directly used for age comparison between different subject groups or cohorts, which can provide richer information than only ``relative relations" proposed in~\cite{lim2019order}.
(2) Only healthy subjects are involved in this study. However, the age differences between the healthy and diseased subjects can be compared by the proposed method with the pair of the input images $(x,y)$.
In the future, we will apply the model to compute the brain age difference when $x$ is from the healthy cohort and $y$ is from the disease cohort. 
(3) As shown in Table~\ref{tab:dataset} and Fig.~\ref{fig:datadistribution}, the age distribution of our dataset is unbalanced, with more samples in 15-30 years and elderly ages.
As a result, we see bigger errors in age ranges where fewer samples are available~\cite{he2021multi}.
Feng et al.~\cite{feng2020estimating} recently showed that balancing the age distribution (by under-sampling in more popular age bins) may reduce the errors in less popular age bins.
However, under-sampling the data created a smaller overall sample size.
Our ongoing work is on gathering more images~\cite{cole2017predicting,feng2020estimating}, and even resampling or synthesizing new images, so that we can increase the sample size while balancing the age distribution.
(4) The proposed deep relation learning can provide good accuracy for brain age estimation. 
However, the deep neural networks are usually hard to be interpreted, especially for the convolutional neural networks which are used to extract deep features.
One future direction is to interpret the Transformer by visualizing the attention heatmaps based on the deep Taylor decomposition principle~\cite{chefer2021transformer}.
These attention heatmaps may be useful to understand the interaction and influence among the four relations for brain age estimation.
(5) Table~\ref{tab:soacomparison} compares the proposed method to other state-of-the-art models with the same training configurations. However, another fair comparison might be using different models at their own optimal parameters. 
In practice, we have found that these parameters do not affect the accuracies if they are in a reasonable range (e.g., learning rate between 0.001 and 0.0001). 
We also ran SFCN~\cite{peng2019accurate}, another 3D model, with only the T1w image and with the parameters specified in its original paper~\cite{peng2019accurate}, and we found a accuracy (MAE=2.67±0.11) statistically equivalent to SFCN’s accuracy we listed in Table~\ref{tab:soacomparison} (MAE=2.62±0.07), which does not affect the ranking and findings in the comparison. In future work, we will more thoroughly study how the training setting affects the strong backbone for deep relation learning.

In conclusion, we have proposed a novel deep relation learning for brain age estimation and our proposed method can achieve lower MAE than the other six state-of-the-art models.
The proposed method was validated on a lifespan dataset with 5-fold cross-validation, yielding an MAE of 2.38 years, CS($\alpha=5$ years) of 87.40\% and Pearson correlation of 0.988.

\section*{Acknowledgment}
This work was funded, in part, by the Harvard Medical School and Boston Children’s Hospital Faculty Development Award (YO),St Baldrick Foundation Scholar Award Grace Fund (YO),  R03 HD104891 (YO), R21 NS121735 (YO), and Charles A. King Trust Research Fellowship (SH).

\bibliographystyle{IEEEtran}
\bibliography{IEEEabrv,BrainAge.bib}

\end{document}